\PassOptionsToPackage{table,xcdraw}{xcolor}
\PassOptionsToPackage{most}{tcolorbox} 
\documentclass[11pt, a4paper, logo, copyright, nonumbering]{wechat}
\usepackage[numbers, square, sort&compress]{natbib}
\usepackage{dblfloatfix}
\usepackage{ulem}
\usepackage{caption}
\definecolor{citecolor}{HTML}{1976D2}
\hypersetup{
    colorlinks=true,
    linkcolor=black,
    filecolor=magenta,
    urlcolor=blue!50!black,
    citecolor=citecolor,
}
\usepackage{dramatist}
\usepackage{xspace}
\usepackage{pifont}
\usepackage{multirow}
\usepackage{tcolorbox}
\usepackage{xltabular}
\usepackage{longtable}
\usepackage{hyperref}
\usepackage{diagbox}
\usepackage{makecell}
\usepackage{amsmath}
\usepackage{amssymb}
\usepackage{amsfonts}
\usepackage{bm}
\usepackage{lineno}

\usepackage[bottom]{footmisc}

\usepackage{CJKutf8}

\usepackage{setspace}
\usepackage{subcaption} 

\usepackage{titlesec}

\usepackage{tcolorbox}
\usepackage{fontawesome5}
\usepackage{enumitem}

\usepackage{booktabs}
\usepackage{multirow}
\usepackage[table,xcdraw]{xcolor}
\usepackage{graphicx}

\usepackage{subcaption}
\usepackage{microtype}
\usepackage{mathtools}
\usepackage{array}
\usepackage{tabularx}
\usepackage{colortbl}
\usepackage{listings}

\usepackage{tikz}
\usetikzlibrary{arrows.meta,positioning,fit,backgrounds}

\usepackage{amsfonts}       
\usepackage{nicefrac}       
\usepackage{microtype}      
\usepackage{xcolor}         
\usepackage{amsmath}
\usepackage{algorithm}
\usepackage{algpseudocode}
\usepackage{bbm}
\usepackage[most]{tcolorbox}
\usepackage{multirow}
\usepackage[table]{xcolor}
\usepackage{enumitem}
\usepackage{wrapfig}
\usepackage{subcaption}
\usepackage{titletoc}
\usepackage[most]{tcolorbox}
\usepackage{xcolor}
\usepackage{graphicx}
\usepackage{amsmath}
\usepackage{amssymb}
\usepackage{enumitem}
\usepackage{booktabs}
\usepackage{placeins}
\usepackage{tabularx}

\definecolor{caseorange}{RGB}{255,210,165}
\definecolor{caseborder}{RGB}{245,170,90}
\definecolor{lightgraybg}{RGB}{248,248,248}
\definecolor{percpurple}{RGB}{122,57,226}
\definecolor{reasorange}{RGB}{235,110,0}
\definecolor{answerred}{RGB}{210,0,0}

\newtcolorbox{samplebox}[1]{
    enhanced,
    breakable,
    colback=lightgraybg,
    colframe=caseborder,
    coltitle=white,
    colbacktitle=caseorange,
    title=\textbf{#1},
    fonttitle=\large\bfseries,
    boxrule=0.8pt,
    arc=2mm,
    left=3mm,
    right=3mm,
    top=2mm,
    bottom=2mm,
    attach boxed title to top left={xshift=0mm,yshift=0mm},
    boxed title style={
        colback=caseorange,
        colframe=caseorange,
        boxrule=0pt,
        arc=2mm,
        left=2mm,
        right=2mm,
        top=1mm,
        bottom=1mm
    }
}

\usepackage[nameinlink,capitalize,noabbrev]{cleveref}

\tcbset{
    contributionbox/.style={
        colback=blue!3,
        colframe=blue!50!black,
        boxrule=0.5pt,
        arc=3pt,
        left=6pt,
        right=6pt,
        top=6pt,
        bottom=6pt,
        fonttitle=\bfseries
    }
}

\titlespacing*{\paragraph}
  {0pt}                   
  {0.5ex plus 1ex minus .2ex} 
  {1em}

\makeatletter
\def\@BTrule[#1]{%
  \ifx\longtable\undefined
    \let\@BTswitch\@BTnormal
  \else\ifx\hline\LT@hline
    \nobreak
    \let\@BTswitch\@BLTrule
  \else
     \let\@BTswitch\@BTnormal
  \fi\fi
  \global\@thisrulewidth=#1\relax
  \ifnum\@thisruleclass=\tw@\vskip\@aboverulesep\else
  \ifnum\@lastruleclass=\z@\vskip\@aboverulesep\else
  \ifnum\@lastruleclass=\@ne\vskip\doublerulesep\fi\fi\fi
  \@BTswitch}
\makeatother

\addto\extrasenglish{
}

 {\begin{list}{}%
         {\setlength{\leftmargin}{#1}}%
         \item[]%
 }
 {\end{list}}

\reportnumber{001} 

\colorlet{failred}{red!75!black}
\colorlet{okgreen}{green!50!black}
\colorlet{obsgray}{black!55}
\newtcolorbox{casebox}[1]{%
  enhanced, breakable, colback=white, colframe=black!55,
  boxrule=0.5pt, arc=1.2mm, left=2.2mm, right=2.2mm, top=1.8mm, bottom=1.8mm,
  title={#1}, coltitle=white, colbacktitle=black!68,
  fonttitle=\footnotesize\bfseries, before skip=8pt, after skip=8pt}
\newtcolorbox{basepanel}{%
  colback=failred!5, colframe=failred!80, boxrule=0.5pt, arc=1mm,
  left=1.6mm, right=1.6mm, top=1.2mm, bottom=1.2mm,
  title={$\times$~\textbf{Base} (Qwen3.5-9B)}, coltitle=white, colbacktitle=failred!85,
  fonttitle=\scriptsize\bfseries}
\newtcolorbox{ourspanel}{%
  colback=okgreen!6, colframe=okgreen!85, boxrule=0.5pt, arc=1mm,
  left=1.6mm, right=1.6mm, top=1.2mm, bottom=1.2mm,
  title={\textbf{Pass: SkillRubric (Ours)}}, coltitle=white, colbacktitle=okgreen!90,
  fonttitle=\scriptsize\bfseries}

\usepackage[most]{tcolorbox}
\usepackage{listings}
\tcbuselibrary{listings,breakable,skins}

\definecolor{prompttitle}{HTML}{287780}
\definecolor{promptborder}{HTML}{A9C8CC}
\definecolor{promptbackground}{HTML}{F7F9F9}

\lstdefinestyle{promptstyle}{
  basicstyle=\small\ttfamily,
  columns=fullflexible,
  keepspaces=true,
  showspaces=false,
  showstringspaces=false,
  breaklines=true,
  breakatwhitespace=true,
  breakautoindent=false,
  breakindent=0pt,
  xleftmargin=0pt,
  framexleftmargin=0pt,
  aboveskip=0pt,
  belowskip=0pt,
  escapeinside={(*@}{@*)}
}

\newtcblisting{promptbox}[1]{%
  enhanced,
  breakable,
  listing only,
  listing engine=listings,
  colback=promptbackground,
  colframe=promptborder,
  boxrule=0.6pt,
  arc=0mm,
  left=4mm,
  right=4mm,
  top=8mm,
  bottom=3mm,
  title={#1},
  colbacktitle=prompttitle,
  coltitle=white,
  fonttitle=\bfseries\ttfamily,
  attach boxed title to top left={xshift=4mm,yshift=-3mm},
  boxed title style={%
    colback=prompttitle,
    colframe=prompttitle,
    boxrule=0pt,
    arc=1mm,
    left=2mm,
    right=2mm,
    top=1mm,
    bottom=1mm
  },
  listing options={style=promptstyle},
  before skip=8pt,
  after skip=8pt
}

\colorlet{guideblue}{blue!65!black}
\colorlet{rubricorange}{orange!85!black}

\newtcolorbox{skillbox}[1]{%
  enhanced,
  colback=white,
  colframe=black!55,
  boxrule=0.5pt,
  arc=1.2mm,
  left=2.2mm,
  right=2.2mm,
  top=1.8mm,
  bottom=1.8mm,
  title={#1},
  coltitle=white,
  colbacktitle=black!68,
  fonttitle=\footnotesize\bfseries,
  fontupper=\footnotesize,
  before upper={%
    \setlength{\parindent}{0pt}%
    \setlength{\parskip}{1.0mm}%
  },
  before skip=0pt,
  after skip=2pt
}

\newtcolorbox{guidancepanel}{%
  enhanced,
  colback=guideblue!5,
  colframe=guideblue!80,
  boxrule=0.5pt,
  arc=1mm,
  left=1.6mm,
  right=1.6mm,
  top=1.2mm,
  bottom=1.2mm,
  title={Actor-facing guidance $g$ (shown to the policy)},
  coltitle=white,
  colbacktitle=guideblue!85,
  fonttitle=\scriptsize\bfseries,
  fontupper=\scriptsize,
  before upper={%
    \setlength{\parindent}{0pt}%
    \setlength{\parskip}{0.9mm}%
  },
  before skip=0pt,
  after skip=0pt
}

\newtcolorbox{rubricpanel}{%
  enhanced,
  colback=rubricorange!6,
  colframe=rubricorange!85,
  boxrule=0.5pt,
  arc=1mm,
  left=1.6mm,
  right=1.6mm,
  top=1.2mm,
  bottom=1.2mm,
  title={Evaluator-facing rubric $\mathcal{C}$ (hidden from the policy)},
  coltitle=white,
  colbacktitle=rubricorange!90,
  fonttitle=\scriptsize\bfseries,
  fontupper=\scriptsize,
  before upper={%
    \setlength{\parindent}{0pt}%
    \setlength{\parskip}{0.9mm}%
  },
  before skip=0pt,
  after skip=0pt
}

\newcounter{srprompt}[section]

\crefname{srprompt}{Prompt}{Prompts}
\Crefname{srprompt}{Prompt}{Prompts}

\newtcblisting{skillrubricprompt}[1]{%
    enhanced,
    breakable,
    listing only,
    colback=gray!3,
    colframe=gray!45,
    boxrule=0.5pt,
    arc=1pt,
    left=1.5mm,
    right=1.5mm,
    top=1mm,
    bottom=1mm,
    title={#1},
    fonttitle=\bfseries\small,
    listing options={
        basicstyle=\ttfamily\scriptsize,
        breaklines=true,
        breakatwhitespace=false,
        columns=fullflexible,
        keepspaces=true,
        showstringspaces=false,
        tabsize=2
    }
}

\title{\centering Structured Role-Aware Policy Optimization for Multimodal Reasoning}

\author{
    {\large\bfseries
    Bingqing Jiang$^{1,3,\ast,\S}$,
    Guoxi Zhang$^{2,\ast}$,
    Jasper Wang$^{3}$,
    Bingning Wang$^{3,\ddagger}$,
    Difan Zou$^{1,\dagger}$
    }\\
    {\normalsize\normalfont
    $^{1}$The University of Hong Kong
    \quad
    $^{2}$Peking University
    \quad
    $^{3}$WeChat, Tencent
    }
}

\renewcommand{\phi}{\varphi}

\renewcommand{\epsilon}{\varepsilon}
\renewcommand{\imath}{\mathrm{i}}

\newlength{\restsubwidth}
\newlength{\restsubheight}
\newlength{\restsubmoreheight}
\newcommand{\rest}[2]{%
        \settowidth{\restsubwidth}{\ensuremath{#2}}
        \settoheight{\restsubheight}{\ensuremath{{}_{#2}}}
        \ensuremath{{#1\hskip 0.5pt}_{\vrule\kern2pt\parbox[b][%
        4pt][b]{\the\restsubwidth}{%
                        \ensuremath{{}_{#2}}}}}
        }

\begin{abstract}
Reinforcement learning from verifiable rewards (RLVR), especially with Group Relative Policy Optimization (GRPO), has shown strong potential for improving the reasoning capabilities of large vision-language models (LVLMs). However, in multimodal reasoning, final-answer rewards are typically assigned at the sequence level and do not distinguish the functional roles of different tokens, making it difficult to determine whether a correct answer is supported by task-relevant visual evidence. In this paper, we revisit multimodal RLVR from the perspective of role-aware token-level credit assignment, where structured responses are decomposed into perception tokens for extracting visual evidence and reasoning tokens for deriving answers from that evidence. Based on this perspective, we propose \textbf{S}tructured \textbf{R}ole-aware \textbf{P}olicy \textbf{O}ptimization (\textbf{SRPO}), which refines the sequence-level GRPO advantage into role-aware token-level advantages without changing the reward function. Specifically, SRPO assigns role-specific credit by using self-distilled on-policy contrasts: perception tokens are emphasized according to their visual dependency under original versus corrupted visual inputs, while reasoning tokens are emphasized according to their consistency with the generated perception. These role-specific signals are further unified through a shared trajectory-level baseline, yielding positive token weights that adjust relative update magnitudes while preserving the original GRPO reward and advantage signs, without requiring external reward models or separate teachers. Experiments across diverse multimodal reasoning benchmarks show that SRPO improves evidence-grounded reasoning, highlighting the importance of moving beyond uniform sequence-level credit toward role-aware optimization for reliable multimodal reasoning.
\end{abstract}

\begin{document}
\begin{CJK*}{UTF8}{gbsn}

\maketitle
\setlength{\parindent}{0pt} 
\enlargethispage{1cm}

\newcommand\blfootnote[1]{%
  \begingroup
  \renewcommand\thefootnote{}\footnote{#1}%
  \addtocounter{footnote}{-1}%
  \endgroup
}


\blfootnote{$^\ast$ These authors contributed equally to this work.}
\blfootnote{$^\dagger$ Corresponding author.}
\blfootnote{$^\ddagger$ Project leader.}
\blfootnote{$^\S$ Work done during an internship at WeChat.}

\section{Introduction}

\begin{figure}[t]
    \centering
    \includegraphics[width=0.6\linewidth]{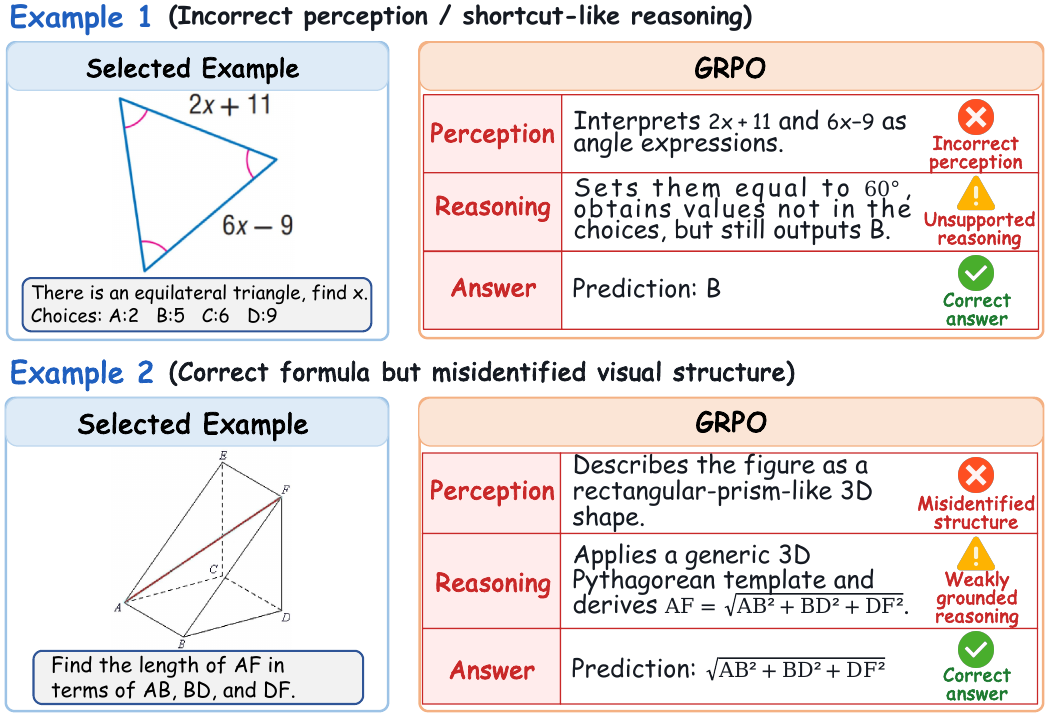}\vspace{-1mm}
    \caption{MathVerse examples where GRPO reaches correct answers despite flawed intermediate grounding.}\vspace{-1mm}
    \label{fig:intro_motivation}
\end{figure}

\begin{table*}[t]
\centering
\footnotesize
\setlength{\tabcolsep}{5pt}
\renewcommand{\arraystretch}{1.10}
\caption{
Motivation diagnostics on Qwen2.5-VL base models.
Original and Mismatch report regenerated-answer accuracy (\%);
CAR is the percentage of originally correct questions that remain
correct after regeneration with a mismatched image.
The last three columns report log-probability decreases
(nats/token) for fixed response tokens:
image masking for perception (P) and reasoning (R),
and removal of P for R.
R retains the original P context during image masking.
}\vspace{-3mm}
\label{tab:visual_diagnostic}
\begin{tabular}{ll|ccc|ccc}
\toprule
\textbf{Benchmark}
& \textbf{Scale}
& \textbf{Original}
& \textbf{Mismatch}
& \textbf{CAR}
& \textbf{Mask image: P}
& \textbf{Mask image: R}
& \textbf{Remove P: R} \\
\midrule
LogicVista
& 3B & 40.04 & 21.77 & 25.33
& 0.2756 & 0.0377 & 0.2746 \\
LogicVista
& 7B & 44.30 & 23.64 & 27.95
& 0.2856 & 0.0453 & 0.1632 \\
\midrule
MathVerse
& 3B & 29.06 & 12.99 & 21.98
& 0.1958 & 0.0087 & 0.1742 \\
MathVerse
& 7B & 42.77 & 15.78 & 22.85
& 0.2127 & 0.0097 & 0.1337 \\
\bottomrule
\end{tabular}
\vspace{-4mm}
\end{table*}

Reinforcement learning from verifiable rewards (RLVR)~\citep{guo2025deepseek}, particularly with online policy optimization algorithms such as Group Relative Policy Optimization (GRPO)~\citep{deepseek-math}, has recently emerged as an effective paradigm for improving the reasoning capabilities of large language models.
Motivated by its success in text-only settings, a growing body of work has begun to extend RLVR to large vision-language models (LVLMs) for multimodal reasoning~\citep{huang2026visionr1,zhang2025r1,yao2025r1,wang2025skywork,zha2026vision}.
Despite encouraging progress, multimodal reasoning remains substantially more challenging than its text-only counterpart: producing a reliable answer requires the model to first identify task-relevant visual evidence and then use that evidence to support subsequent reasoning~\citep{liu2026noisyrollout,wang2026vlrethinker,huang2026spotlight,miao2026seeing}.
For evidence-dependent multimodal reasoning, sequence-level answer supervision has a central limitation: it verifies only whether the final response reaches the correct answer, without assessing whether the intermediate tokens are grounded in task-relevant visual evidence~\citep{li2025vision,wang2026perceptionaware,chen2025perception}.
In practice, LVLMs can produce correct or plausible answers by exploiting language priors, dataset regularities, or shallow textual heuristics, rather than faithfully grounding their responses in the image~\citep{ghosh2025visual,fang2026grounding}.
As illustrated in Fig.~\ref{fig:intro_motivation}, our MathVerse~\citep{zhang2024mathverse} evaluation shows that GRPO can produce correct answers despite flawed intermediate grounding.
In the first example, the model mistakes side-length expressions for angles and arrives at the correct answer through shortcut-like reasoning.
In the second, it derives the correct formula despite misidentifying the 3D structure.
These rollouts may still receive positive sequence-level rewards despite their grounding errors, highlighting that higher answer accuracy does not necessarily imply better image-grounded reasoning.

Recent efforts have begun to address this limitation by separating
visual perception from downstream reasoning~\citep{huang2026spotlight,miao2026seeing,wang2026perceptionaware,zhou2025proreason}.
A structured response makes these functions explicit:
a \emph{perception} segment extracts question-relevant visual evidence,
and a \emph{reasoning} segment uses that evidence to derive
an answer~\citep{sharma2026see}.
However, structure alone does not determine how learning credit
should be distributed between the two segments.
This raises a concrete question:
\emph{do perception and reasoning tokens exhibit different
information dependencies that warrant different credit signals?}

To examine how visual evidence supports perception and reasoning,
we conduct controlled diagnostics on the original Qwen2.5-VL-3B
and Qwen2.5-VL-7B models using LogicVista and
MathVerse in Table~\ref{tab:visual_diagnostic}.
We first replace the original image with a mismatched one and
regenerate the response.
Accuracy drops by 16.07--26.99 percentage points, demonstrating
substantial dependence on matched visual input, while
21.98--27.95\% of originally correct questions remain correct,
as measured by the correctness retention rate (CAR).
To further localize these information dependencies within the
original responses, we hold each response fixed and rescore
its tokens under two context interventions: masking the image
while retaining the textual context, and removing the perception
segment while retaining the image and question.
We measure sensitivity as the average decrease in token
log-probability, expressed in nats per token.
Across all four settings, image masking decreases perception-token
log-probabilities more than reasoning-token log-probabilities,
with a gap of 0.1871--0.2403 nats/token.
Removing the perception segment, in turn, decreases reasoning-token
log-probabilities by 0.1337--0.2746 nats/token.
Together, these results support distinct information dependencies:
perception tokens are more directly sensitive to the image,
whereas reasoning tokens draw conditional support from the generated
perception, which remains available during image masking.
These findings motivate a role-aware approach to credit assignment
that reflects how each segment contributes to visually grounded
answer generation.

To account for the distinct information dependencies of perception and reasoning, we propose \textbf{SRPO} (\textbf{S}tructured \textbf{R}ole-aware \textbf{P}olicy \textbf{O}ptimization), a token-level policy optimization method for multimodal RLVR.
SRPO structures each response into a \emph{perception} segment that extracts task-relevant visual evidence and a \emph{reasoning} segment that derives the answer from it, and converts the sequence-level GRPO advantage into role-aware token-level advantages.
Its positive modulation weights follow a \textit{self-distilled credit assignment} principle: the frozen rollout policy rescores the generated tokens under role-specific information conditions.
To reflect perception's direct image dependence, SRPO contrasts token likelihoods under original and corrupted images.
For reasoning, motivated by its dependence on the generated perception context, SRPO assesses grounding consistency by contrasting perception-conditioned prediction with direct image--question grounding.
These scores are centered using a shared response-level baseline over the valid trajectory and mapped to bounded positive multipliers. The resulting token-level advantages retain the sign of the original GRPO advantage, without changing the verifiable reward or requiring external reward models, auxiliary annotations, or separate teachers.
In summary, our contributions are threefold:
\begin{itemize}[leftmargin=*]
\item Compared with conventional sequence-level optimization that assigns a uniform credit signal to the whole response, we introduce a \emph{role-decomposed perspective} for multimodal token-level credit assignment. This perspective treats multimodal responses as functionally heterogeneous trajectories, where the \emph{perception} stage extracts task-relevant visual evidence and the \emph{reasoning} stage derives answers from that evidence.
\item We propose SRPO, which instantiates role-decomposed token-level credit assignment within GRPO.
SRPO estimates perception- and reasoning-stage credit through role-specific self-distilled on-policy contrasts, then calibrates these signals with a shared trajectory-level baseline and converts them into bounded positive weights to reweight the sequence-level GRPO advantage.
This preserves the original reward and the sign of each token's advantage without external reward models or separate teachers.
\item We demonstrate that SRPO improves evidence-grounded multimodal reasoning across multiple benchmarks, with ablations confirming the importance of role-specific credit signals and unified trajectory-level modulation.
\end{itemize}

\vspace{-2mm}
\section{Related Work}
\vspace{-2mm}

\paragraph{Multimodal Reinforcement Learning for Reasoning.}
RLVR with algorithms such as PPO and GRPO improves reasoning
in large language models~\citep{guo2025deepseek,deepseek-math},
motivating RL-style post-training for
LVLMs~\citep{achiam2023gpt,Qwen2.5-VL,2024arXiv240305530G,zhu2025internvl3}.
Existing approaches advance multimodal reasoning through
improved training data~\citep{huang2026visionr1,yao2025r1},
rollout diversification~\citep{liu2026noisyrollout,wang2026vlrethinker},
replay mechanisms~\citep{wang2025skywork},
and optimization heuristics~\citep{zha2026vision}.

\vspace{-3mm}
\paragraph{Perception-Aware and Structured Multimodal Reasoning.}
Correct answers do not necessarily imply visually grounded
reasoning~\citep{ghosh2025visual}.
Prior work strengthens grounding through perception-aware
rewards~\citep{xia2025visionary,xiao2025advancing,wang2026perceptionaware},
explicit perceptual
scaffolds~\citep{cao2025ground,sarch2026grounded,su2026pixel,sun2026regionreasoner},
and RL mechanisms that encourage reliance on visual
evidence~\citep{yu2026perceptionr,wang2026visually,liu2026visionreasoner}.
Related approaches structure generation into perception,
evidence extraction, reasoning, and answer
prediction~\citep{chen2025perception,li2026visionsr,miao2026seeing,ding2025vtperception,he2025visplay}.
SRPO uses explicit perception--reasoning structure to guide
token-level policy updates, linking each segment's functional
role to its credit assignment.

\vspace{-3mm}
\paragraph{Process Supervision and Fine-Grained Credit Assignment.}
Process reward models (PRMs) use human or automated step-level supervision for response selection and policy optimization~\citep{lightman2023lets,wang2024mathshepherd}.
VisualPRM extends this to multimodal best-of-$N$ selection~\citep{wang2025visualprm}.
Related work studies segment- or token-level credit in text reasoning~\citep{guo2026segment,tran2025exploiting,wang2026beyond,xu2025kdrl,zhang2026reinforcement,yang2026self},
and token reweighting and perception-aware optimization in multimodal reasoning~\citep{huang2026spotlight,lu2026bridging,ye2026not}.
SRPO assigns token-level credit based on image dependence for perception and perception-context dependence for reasoning, retaining GRPO's verifiable outcome reward without a separately trained PRM.

\begin{figure*}[t]
    \centering
    \includegraphics[width=\linewidth]{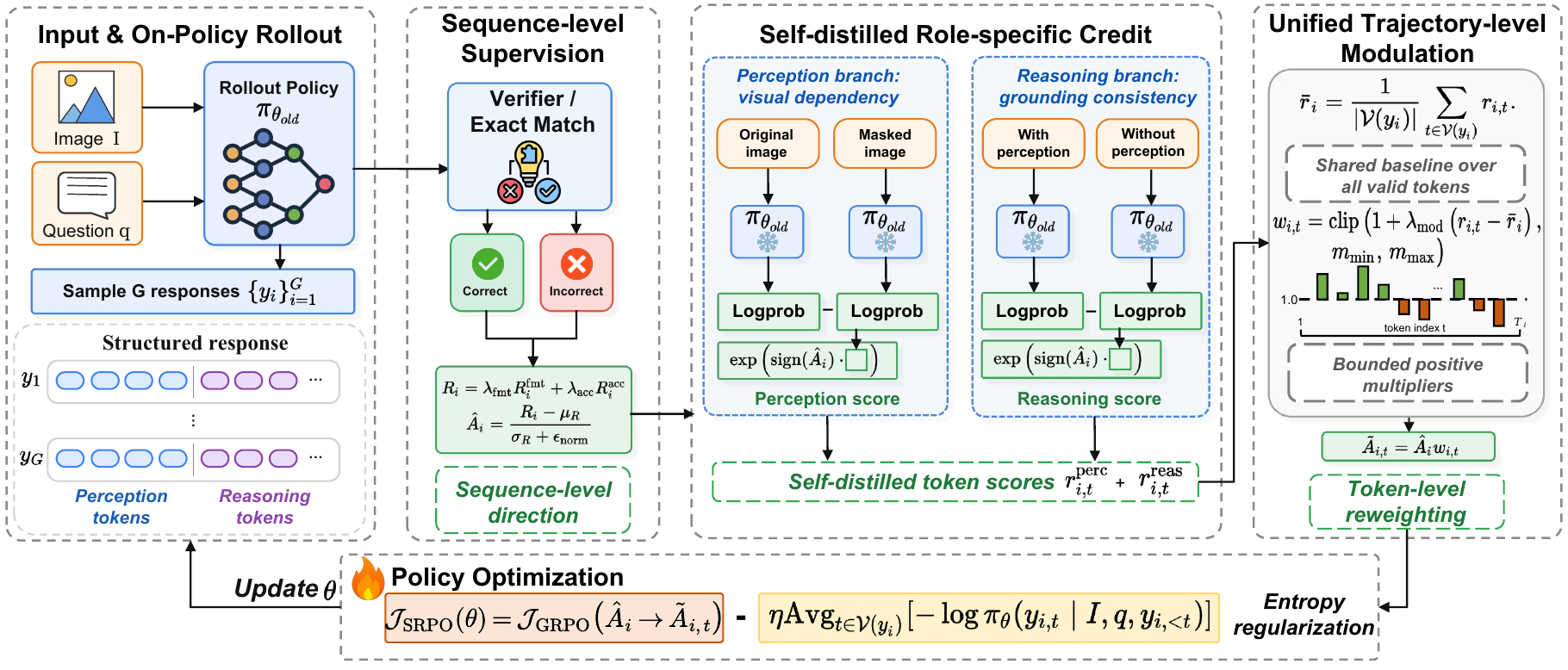}\vspace{-2mm}
    \caption{\textbf{Overview of SRPO}.
        SRPO decomposes each on-policy response into perception and reasoning tokens, obtains a sequence-level advantage from verifier rewards, and converts it into role-aware token-level advantages.
        Perception tokens are scored by visual dependency, while reasoning tokens are scored by perception-supported grounding consistency.
        The resulting scores produce bounded positive modulation weights for token-level reweighting, and policy optimization further includes a response-token entropy regularizer for stability.}\vspace{-5mm}
    \label{fig:method_overview}
\end{figure*}

\vspace{-2mm}
\section{Method}
\label{sec:method}
\vspace{-2mm}

\subsection{Preliminary: GRPO} 
\label{sec:prelim_grpo} 
\vspace{-1mm}

GRPO is a value-free reinforcement learning algorithm that estimates policy-gradient advantages by comparing multiple responses sampled from the same prompt~\citep{deepseek-math}.
Given an image--question pair $(I,q)$, the behavior policy $\pi_{\theta_{\mathrm{old}}}$ samples a group of $G$ responses $
\{y_i\}_{i=1}^{G}, 
y_i = (y_{i,1},\ldots,y_{i,T_i})
\sim \pi_{\theta_{\mathrm{old}}}(\cdot \mid I,q)$.
A verifier assigns each complete response a scalar reward that combines format compliance and answer correctness:
\[
R_i
=
\lambda_{\mathrm{fmt}} R_i^{\mathrm{fmt}}
+
\lambda_{\mathrm{acc}} R_i^{\mathrm{acc}},
\]
where $R_i^{\mathrm{fmt}}$ measures whether the response follows the required output format, and
$
R_i^{\mathrm{acc}}
=
\mathbbm{1}\{\mathrm{eq}(\hat a_i,a)\}$
measures whether the extracted answer $\hat a_i$ matches the ground-truth answer $a$ under an equivalence checker $\mathrm{eq}(\cdot,\cdot)$.
GRPO estimates the advantage of each response by normalizing rewards within the group as $\hat A_i=(R_i-\mu_R)/(\sigma_R+\epsilon_{\mathrm{norm}})$, where $\mu_R=\frac{1}{G}\sum_{j=1}^{G}R_j$ and $\sigma_R=\sqrt{\frac{1}{G}\sum_{j=1}^{G}(R_j-\mu_R)^2}$ are the group reward mean and standard deviation, respectively, and $\epsilon_{\mathrm{norm}}$ is a small constant for numerical stability.
The policy is then optimized with a PPO-style clipped surrogate objective:
\[
\begin{aligned}
\mathcal{J}_{\mathrm{GRPO}}(\theta)
= \mathbb{E}\Biggl[
\frac{1}{G}\sum_{i=1}^{G}
\frac{1}{T_i}\sum_{t=1}^{T_i}
\min\Bigl(\rho_{i,t}(\theta)\hat A_i,
\mathrm{clip}\bigl(
\rho_{i,t}(\theta),1-\epsilon,1+\epsilon
\bigr)\hat A_i
\Bigr)\Biggr].
\end{aligned}
\]
where
$
\rho_{i,t}(\theta)
\!=\!
\frac{
\pi_{\theta}(y_{i,t}\mid I,q,y_{i,<t})
}{
\pi_{\theta_{\mathrm{old}}}(y_{i,t}\mid I,q,y_{i,<t})
}$.
In practice, a KL penalty to a reference policy $\pi_{\mathrm{ref}}$ can be added to constrain policy drift.
Despite its effectiveness, GRPO performs credit assignment at the sequence level: the same scalar advantage $\hat A_i$ is applied uniformly to all valid tokens in a response.
This coarse-grained design ignores the functional heterogeneity of multimodal reasoning trajectories.
In particular, perception tokens should be encouraged according to their dependence on visual evidence, whereas reasoning tokens should be evaluated by whether the subsequent inference remains supported by the perceived content.
A uniform sequence-level advantage cannot distinguish these roles.
This motivates a structured token-level modulation mechanism that preserves the sign of the sequence-level advantage while redistributing its magnitude across tokens.

\vspace{-1mm}
\subsection{Structured Role-aware Policy Optimization}
\label{sec:srca}
\vspace{-1mm}

Motivated by the role-dependent information sensitivities in Table~\ref{tab:visual_diagnostic}, we introduce SRPO, which combines role-specific credit estimation with
shared trajectory-level modulation.
Perception and reasoning tokens receive distinct
conditional-dependence scores, but both sets of scores
are calibrated using a common response-level baseline
before being converted into bounded positive weights.
We next describe the structured response format,
the role-specific scoring functions,
and their unified modulation of the GRPO advantage.
For each on-policy rollout batch, SRPO rescores the generated tokens with the behavior policy $\pi_{\theta_{\mathrm{old}}}$ under controlled changes of visual or textual context.
These rescoring computations are treated as stop-gradient signals and are used only for credit modulation.
For each sampled response $y_i=(y_{i,1},\ldots,y_{i,T_i})$, we assume a structured format that can be parsed into two contiguous spans $y_i = (y_i^{\mathrm{perc}}, y_i^{\mathrm{reas}})$,
where $y_i^{\mathrm{perc}}$ denotes perception tokens that describe or extract task-relevant visual evidence, and $y_i^{\mathrm{reas}}$ denotes reasoning tokens that use the perceived content to derive the final answer.
SRPO then derives role-specific token scores for the two spans and converts them into token-level modulation weights for the GRPO advantage.

\vspace{-2mm}
\paragraph{Self-distilled visual dependency for perception tokens.}
For perception tokens, we estimate visual reliance by comparing the behavior policy under two visual conditions. The first uses the original image $I$, and the second uses a corrupted image $I^{\mathrm{mask}}$ obtained by masking a subset of image patches. Model parameters are identical across the two evaluations; only the visual input is changed. For each perception-token position $t\in\mathcal{P}_i$, we first compute the stop-gradient log-likelihood contrast
\begin{equation}
\label{eq:perc_delta}
\begin{aligned}
\delta_{i,t}^{\mathrm{perc}}
= \operatorname{sg}\Bigl[
 \log \pi_{\theta_{\mathrm{old}}}
\bigl(y_{i,t}\mid I,q,y_{i,<t}\bigr) 
 {}- \log \pi_{\theta_{\mathrm{old}}}
\bigl(y_{i,t}\mid I^{\mathrm{mask}},q,y_{i,<t}\bigr)
\Bigr].
\end{aligned}
\end{equation}
A positive $\delta_{i,t}^{\mathrm{perc}}$ indicates that the target token is more likely under the original image than under the masked image, whereas a negative value indicates the opposite.
Under proportional perception-credit allocation, SRPO converts this contrast into positive perception scores:
\begin{equation}
\label{eq:perc_prop_weight}
r_{i,t}^{\mathrm{perc}}
=
\exp\left(
\operatorname{sign}(\hat A_i)\cdot \delta_{i,t}^{\mathrm{perc}}
\right),
\quad t\in\mathcal{P}_i.
\end{equation}
When $\hat A_i>0$, visually dependent perception tokens receive larger modulation scores and are therefore reinforced more strongly. 
When $\hat A_i<0$, the sign reversal assigns smaller scores to visually dependent perception tokens and larger scores to weakly visual-dependent ones, so failed trajectories mainly suppress perception tokens that are insufficiently tied to the image.
The stop-gradient operation ensures that these self-distilled signals are used only for credit modulation and do not introduce an auxiliary optimization objective.

\vspace{-2mm}
\paragraph{Self-distilled grounding consistency for reasoning tokens.}
For reasoning tokens, the desired property is not direct visual sensitivity, but whether the generated perception provides useful support for subsequent inference.
We therefore rescore the same on-policy reasoning tokens under the full and perception-ablated contexts, without generating new tokens.
The first is the full rollout context, which contains the image, question, generated perception, and previous reasoning tokens.
The second removes the generated perception block and its corresponding instruction, while retaining the image, question content, and the same reasoning-prefix tokens.
This comparison estimates the support provided by the full perception context relative to the perception-ablated context.
For each reasoning token position $t\in\mathcal{R}_i$, we define
\begin{equation}
\label{eq:reas_delta}
\begin{aligned}
\delta_{i,t}^{\mathrm{reas}}
= \operatorname{sg}\Bigl[
 \log \pi_{\theta_{\mathrm{old}}}
\bigl(y_{i,t}\mid I,q,y_i^{\mathrm{perc}},
y_{i,<t}^{\mathrm{reas}}\bigr) 
 {}- \log \pi_{\theta_{\mathrm{old}}}
\bigl(y_{i,t}\mid I,q,y_{i,<t}^{\mathrm{reas}}\bigr)
\Bigr].
\end{aligned}
\end{equation}
where $y_{i,<t}^{\mathrm{reas}}$ denotes the reasoning-prefix tokens before position $t$.
Here, $q$ includes the corresponding branch-specific instruction given in Appendix~\ref{app:prompt_templates}.
This residual measures the likelihood difference between the full and perception-ablated contexts.
A positive $\delta_{i,t}^{\mathrm{reas}}$ indicates that the reasoning token is more likely under the full context than under the perception-ablated context, whereas a negative value indicates the opposite.
As in perception-token modulation, we use this self-distilled residual in an outcome-aware manner by interpreting it under the sign of the trajectory-level advantage and mapping it to a positive modulation factor:
\begin{equation}
\label{eq:reas_modulation}
r_{i,t}^{\mathrm{reas}} 
=
\exp\left(
\operatorname{sign}(\hat A_i)\cdot \delta_{i,t}^{\mathrm{reas}}
\right),
\quad t\in\mathcal{R}_i .
\end{equation}
When $\hat A_i>0$, perception-supported reasoning tokens with larger $\delta_{i,t}^{\mathrm{reas}}$ receive larger modulation factors, thereby reinforcing successful inference patterns that are grounded in the generated perception.
When $\hat A_i<0$, the sign reversal assigns smaller
modulation factors to perception-supported reasoning
tokens, reducing their penalty in the policy surrogate.

\definecolor{ourscolor}{RGB}{255,239,219}
\definecolor{blockcolor}{RGB}{238,241,247}

\begin{table*}[t]
\centering
\small
\setlength{\tabcolsep}{3.8pt}
\renewcommand{\arraystretch}{1.08}
\caption{
Main results on nine multimodal reasoning benchmarks with
Qwen2.5-VL-3B, Qwen2.5-VL-7B, and Qwen3-VL-8B backbones.
We report scores averaged over four repeated greedy evaluations
of each checkpoint and their unweighted average across benchmarks.
The best and second-best results within each backbone are
highlighted in bold and underlined, respectively.
}\vspace{-2mm}
\label{tab:main_results}
\resizebox{\linewidth}{!}{
\begin{tabular}{l|ccccc|cccc|c}
\toprule
\multirow{2}{*}{\textbf{Model}}
& \multicolumn{5}{c|}{\textbf{Math-Related}}
& \multicolumn{4}{c|}{\textbf{General Task}}
& \multirow{2}{*}{\textbf{Avg.}} \\
\cmidrule(lr){2-6}
\cmidrule(lr){7-10}
& \textbf{MathVista}
& \textbf{MathVision}
& \textbf{We-Math}
& \textbf{MathVerse}
& \textbf{DynaMath}
& \textbf{MMMU-Pro}
& \textbf{MM-Vet}
& \textbf{LogicVista}
& \textbf{NaturalBench}
& \\
\midrule

\rowcolor{blockcolor}
\multicolumn{11}{c}{\textit{\textbf{Based on Qwen2.5-VL-3B}}} \\
\midrule
Vision-SR1-3B
& 64.30 & 27.71 & 26.98 & 30.04 & 45.09
& \underline{30.35} & 60.45 & 39.93 & 63.76 & 43.18 \\
PAPO-G-3B
& 65.35 & 26.40 & 33.57 & 32.61 & 44.19
& 27.92 & \underline{60.50} & 43.79 & \underline{74.68} & 45.45 \\
PAPO-D-3B
& 66.53 & 26.40 & 35.62 & 33.50 & 46.17
& 28.08 & \textbf{60.86} & 43.57 & 72.75 & \underline{45.94} \\
PRCO-3B
& \underline{68.23} & 27.02 & \textbf{37.21}
& \underline{36.14} & \underline{48.44}
& 26.42 & 57.96 & 44.46 & 67.05 & 45.88 \\
\midrule
Qwen2.5-VL-3B
& 63.70 & 23.85 & 23.88 & 30.14 & 43.25
& 25.87 & 54.15 & 41.05 & 74.12 & 42.22 \\
+ GRPO
& 64.85 & \underline{28.12} & 28.38 & 32.30 & 45.40
& 26.11 & 55.37 & 43.01 & 63.58 & 43.01 \\
+ DAPO
& 67.23 & 26.73 & 31.71 & 32.74 & 44.53
& 26.85 & 56.26 & \underline{44.63} & 63.71 & 43.82 \\
\rowcolor{ourscolor}
\textbf{+ SRPO}
& \textbf{69.27} & \textbf{28.70} & \underline{36.15}
& \textbf{38.10} & \textbf{50.06}
& \textbf{30.87} & 59.41 & \textbf{45.75}
& \textbf{74.81} & \textbf{48.12} \\

\midrule
\rowcolor{blockcolor}
\multicolumn{11}{c}{\textit{\textbf{Based on Qwen2.5-VL-7B}}} \\
\midrule
Vision-SR1-7B
& 71.07 & \underline{30.35} & 34.08 & 32.77 & 49.57
& 34.84 & 65.09 & 49.55 & 77.50 & 49.42 \\
ThinkLite-VL-7B
& 73.88 & 26.23 & 38.95 & 39.05 & 53.45
& 35.69 & 64.87 & 45.08 & 77.81 & 50.56 \\
PAPO-G-7B
& 72.82 & 29.61 & 43.62 & 40.80 & 53.97
& 36.34 & 64.31 & 44.35 & \underline{78.22} & 51.56 \\
PAPO-D-7B
& 75.75 & 29.93 & 42.72 & 36.14 & 56.20
& 37.02 & \underline{66.95} & \underline{49.83} & 77.85 & 52.49 \\
Perception-R1-7B
& 73.42 & 28.29 & \textbf{46.88} & 39.31 & 55.64
& 34.49 & 64.56 & 45.41 & 77.85 & 51.76 \\
Vision-Matters-7B
& 72.75 & 30.02 & 39.83 & 40.99 & 55.53
& 36.55 & 64.76 & 49.33 & 77.92 & 51.96 \\
VL-Rethinker-7B
& 73.55 & 29.28 & 43.14 & \underline{42.64} & 55.77
& 37.46 & 64.25 & 44.74 & 78.04 & 52.10 \\
VPPO-7B
& 75.58 & 30.10 & 44.41 & 40.04 & \underline{58.45}
& 38.54 & 65.88 & 48.77 & 78.13 & 53.32 \\
PRCO-7B
& \textbf{76.53} & 27.96 & \underline{46.31} & 40.64 & 57.68
& \underline{39.67} & 66.83 & 48.66 & 78.03 & \underline{53.59} \\
\midrule
Qwen2.5-VL-7B
& 69.42 & 28.37 & 35.00 & 38.71 & 52.41
& 35.12 & 63.60 & 45.97 & 78.11 & 49.63 \\
+ GRPO
& 72.88 & 25.74 & 42.33 & 36.33 & 54.69
& 34.58 & 64.42 & 48.49 & 77.87 & 50.81 \\
+ DAPO
& 74.88 & 29.61 & 42.41 & 37.69 & 58.05
& 34.47 & 66.02 & 48.99 & 78.12 & 52.25 \\
\rowcolor{ourscolor}
\textbf{+ SRPO}
& \underline{75.90} & \textbf{32.69} & 45.36
& \textbf{44.00} & \textbf{58.70}
& \textbf{39.84} & \textbf{68.21} & \textbf{50.95}
& \textbf{78.96} & \textbf{54.96} \\

\midrule
\rowcolor{blockcolor}
\multicolumn{11}{c}{\textit{\textbf{Based on Qwen3-VL-8B}}} \\
\midrule
PAPO-G-8B
& 79.08 & 43.26 & 53.99 & 33.85 & 66.37
& 39.78 & 64.85 & 60.79 & \textbf{80.84} & 58.09 \\
PAPO-D-8B
& \underline{79.35} & 44.33 & 54.79 & 36.07 & 67.26
& 42.20 & 62.66 & 59.96 & \underline{80.66} & 58.59 \\
VPPO-8B
& \textbf{80.33} & \textbf{48.36} & 57.07
& \underline{42.16} & 67.32
& \underline{43.40} & 63.03 & 62.42
& 80.09 & \underline{60.46} \\
PRCO-8B
& 78.53 & 47.04 & \underline{59.14}
& 37.91 & \textbf{68.67}
& 38.14 & 65.63 & \underline{64.21} & 80.09 & 59.93 \\
\midrule
Qwen3-VL-8B
& 76.53 & 43.09 & 52.60 & 35.47 & 66.24
& 42.14 & 65.85 & 54.87 & 80.19 & 57.44 \\
+ GRPO
& 76.75 & 45.15 & 55.43 & 35.25 & 66.50
& 40.51 & 66.11 & 56.38 & 80.09 & 58.02 \\
+ DAPO
& 77.53 & 46.46 & 57.86 & 36.99 & 67.41
& 36.94 & \underline{66.59} & 62.19 & 80.47 & 59.16 \\
\rowcolor{ourscolor}
\textbf{+ SRPO}
& 79.16 & \underline{47.62} & \textbf{59.69}
& \textbf{46.92} & \underline{67.96}
& \textbf{46.48} & \textbf{67.40} & \textbf{64.65}
& \textbf{80.84} & \textbf{62.30} \\
\bottomrule
\end{tabular}
}\vspace{-3mm}
\end{table*}

\vspace{-3mm}
\paragraph{Unified trajectory-level modulation.}
We combine the role-specific raw scores into a single valid-token sequence:
\[
r_{i,t}
=
\begin{cases}
r_{i,t}^{\mathrm{perc}}, & t\in \mathcal{P}_i,\\
r_{i,t}^{\mathrm{reas}}, & t\in \mathcal{R}_i,
\end{cases}
\]
where $\mathcal{P}_i$ and $\mathcal{R}_i$ denote the perception and reasoning token-position sets of response $y_i$, respectively.
Let $\mathcal{V}(y_i)$ denote the set of all valid token positions.
SRPO computes one response-level baseline over the entire valid trajectory:
\[
\bar r_i =
\frac{1}{|\mathcal{V}(y_i)|}
\sum_{t\in\mathcal{V}(y_i)} r_{i,t}.
\]
This baseline is shared by perception and reasoning tokens, rather than computed separately per span.
The centered raw scores are then converted into bounded positive multipliers:
\begin{equation}
\label{eq:simple_mod}
w_{i,t}
\!\!=\!\!
\mathrm{clip}
\left(
1\!\!+\!\!
\lambda_{\mathrm{mod}}
\left(
r_{i,t}\!\!-\!\!\bar r_i
\right),
m_{\min},
m_{\max}
\right),
\quad \!\!\! \!\!\!t \!\!\in\!\! \mathcal{V}(y_i),
\end{equation}
where $\lambda_{\mathrm{mod}}$ is a modulation-strength hyperparameter and $[m_{\min},m_{\max}]$ keeps the multipliers bounded and positive.
The centering step makes each multiplier depend on the token's relative credit within the same response, and the clipping range is chosen such that $w_{i,t}>0$.
The token-level advantage is then defined as
\[
\tilde A_{i,t} = \hat A_i w_{i,t}.
\]
Since $w_{i,t}>0$, the reweighted advantage $\tilde A_{i,t}$ has the same sign as $\hat A_i$ at every valid token. The modulation therefore changes the relative token contributions to the policy-gradient surrogate without reversing their advantage signs.

\vspace{-2mm}
\subsection{Optimization Objective}
\vspace{-2mm}

SRPO retains the form of the GRPO clipped surrogate,
but replaces the uniform sequence-level advantage with
the token-level advantage
$\tilde A_{i,t}=\hat A_i w_{i,t}$:
\[
\begin{aligned}
\mathcal{J}_{\mathrm{SRPO}}(\theta)
= \mathbb{E}\Biggl[
\frac{1}{G}\sum_{i=1}^{G}
\frac{1}{T_i}\sum_{t=1}^{T_i}
\min\Bigl(\rho_{i,t}(\theta)\tilde A_{i,t},
\mathrm{clip}\bigl(
\rho_{i,t}(\theta),1-\epsilon_{\mathrm{low}},
1+\epsilon_{\mathrm{high}}
\bigr)\tilde A_{i,t}
\Bigr)\Biggr],
\end{aligned}
\]
where $\rho_{i,t}(\theta)$ is the PPO importance ratio defined in Section~\ref{sec:prelim_grpo}.
The modulation weights $w_{i,t}$ are computed from stop-gradient contrasts under $\pi_{\theta_{\mathrm{old}}}$ and are fixed during the policy update.
Therefore, SRPO preserves the GRPO optimization form while introducing role-aware token-level reweighting.
However, redistributing update magnitudes across response tokens may amplify updates on unstable or low-confidence tokens in multimodal reasoning trajectories, increasing rollout uncertainty and degrading training stability.
To mitigate this issue, we introduce a lightweight response-token uncertainty penalty along the sampled trajectory.
This term is used only in the final optimization objective and preserves the stop-gradient credit scores and modulation weights.
Concretely, for each sampled response, we define the response-token uncertainty surrogate as
\[
\mathcal{J}_{\mathrm{resp}}(\theta)
=
\mathbb{E}
\left[
\frac{1}{G}
\sum_{i=1}^{G}
\frac{1}{|\mathcal{V}(y_i)|}
\sum_{t\in\mathcal{V}(y_i)}
-\log \pi_\theta
\left(
y_{i,t} \mid  I,q,y_{i,<t}
\right)
\right].
\]
The final regularized objective is defined as
\begin{equation}
\label{eq:srca_total_objective}
\mathcal{J}_{\mathrm{total}}(\theta)
=
\mathcal{J}_{\mathrm{SRPO}}(\theta)
-
\eta\,\mathcal{J}_{\mathrm{resp}}(\theta),
\end{equation}
where $\eta$ controls the strength of the uncertainty penalty.
We maximize $\mathcal{J}_{\mathrm{total}}(\theta)$ during policy optimization.
See Appendix~\ref{app:training_procedure} for the full training procedure.

\vspace{-2mm}
\section{Experiments}
\label{sec:experiments}

\subsection{Experimental Setup}

\paragraph{Models, Data, and Baselines.}
We evaluate SRPO on the Instruct variants of Qwen2.5-VL-3B, Qwen2.5-VL-7B~\citep{Qwen2.5-VL}, and Qwen3-VL-8B~\citep{Qwen3-VL}.
All three models are trained on ViRL39K~\citep{wang2026vlrethinker}, a collection of approximately 39K verifiable multimodal reasoning problems covering diverse visual inputs such as diagrams, charts, and natural images.
We compare against recent open-source multimodal reasoning models at comparable scales, including Vision-SR1~\citep{li2026visionsr}, ThinkLite-VL, PAPO~\citep{wang2026perceptionaware}, Perception-R1~\citep{xiao2025advancing}, Vision-Matters~\citep{li2025vision}, VL-Rethinker~\citep{wang2026vlrethinker}, VPPO~\citep{huang2026spotlight}, and PRCO~\citep{miao2026seeing}.
We re-evaluate publicly released baseline checkpoints, except for PAPO-G and PAPO-D on Qwen3-VL-8B, which we reproduce using their original training prompts.
For controlled comparisons, we train GRPO~\citep{deepseek-math} and DAPO~\citep{yu2026dapo} with the same backbones, data, structured prompts, base rewards, and rollout budgets as SRPO, while retaining method-specific optimization settings.

\vspace{-2mm}
\paragraph{Training Details.}

We implement SRPO and the controlled GRPO and DAPO baselines in EasyR1~\citep{zheng2025easyr1}, using AdamW~\citep{loshchilov2017decoupled} with a learning rate of $1\times10^{-6}$.
We train Qwen2.5-VL models for two epochs and Qwen3-VL-8B for one epoch.
The rollout batch size is 384, with 8 responses sampled per prompt at temperature 1.0 and top-$p$ 0.99.
The maximum response length is 2,048 tokens for Qwen2.5-VL and 4,096 for Qwen3-VL.
The shared base reward weights answer correctness and format compliance by 0.95 and 0.05, respectively.
The training procedure is described in Appendix~\ref{app:training_procedure}, with optimization settings and SRPO hyperparameters detailed in Appendix~\ref{app:training_details}.

\vspace{-3mm}
\paragraph{Evaluation Benchmarks.}
We evaluate on nine multimodal reasoning benchmarks covering math-related and general multimodal reasoning tasks.
The math-related benchmarks include MathVista~\citep{lu2024mathvista}, MathVision~\citep{wang2024measuring}, We-Math~\citep{qiao2024we}, MathVerse~\citep{zhang2024mathverse}, and DynaMath~\citep{zou2025dynamath}.
The general-task benchmarks include MMMU-Pro~\citep{2024arXiv240902813Y}, MM-Vet~\citep{yu2024mm}, LogicVista~\citep{xiao2024logicvistamultimodalllmlogical}, and NaturalBench~\citep{li2024naturalbench}.
We evaluate all methods using VLMEvalKit~\citep{duan2024vlmevalkit} with single-response greedy decoding and report benchmark scores averaged over four evaluation runs.
The overall score is the unweighted mean across all nine benchmarks.
GRPO, DAPO, and SRPO share the same structured evaluation prompt, while external baselines use their published evaluation prompts and the original Qwen models use default benchmark prompts.
For model-based answer evaluation, we use DeepSeek-v4-flash.
Benchmark variants, decoding parameters, and scoring
protocols are detailed in
Appendix~\ref{app:evaluation_details}, while the structured
rollout, evaluation, and auxiliary rescoring templates
are provided in Appendix~\ref{app:prompt_templates}.


\subsection{Main Results}

\begin{figure}[t]
    \centering
    \includegraphics[width=1\textwidth]
    {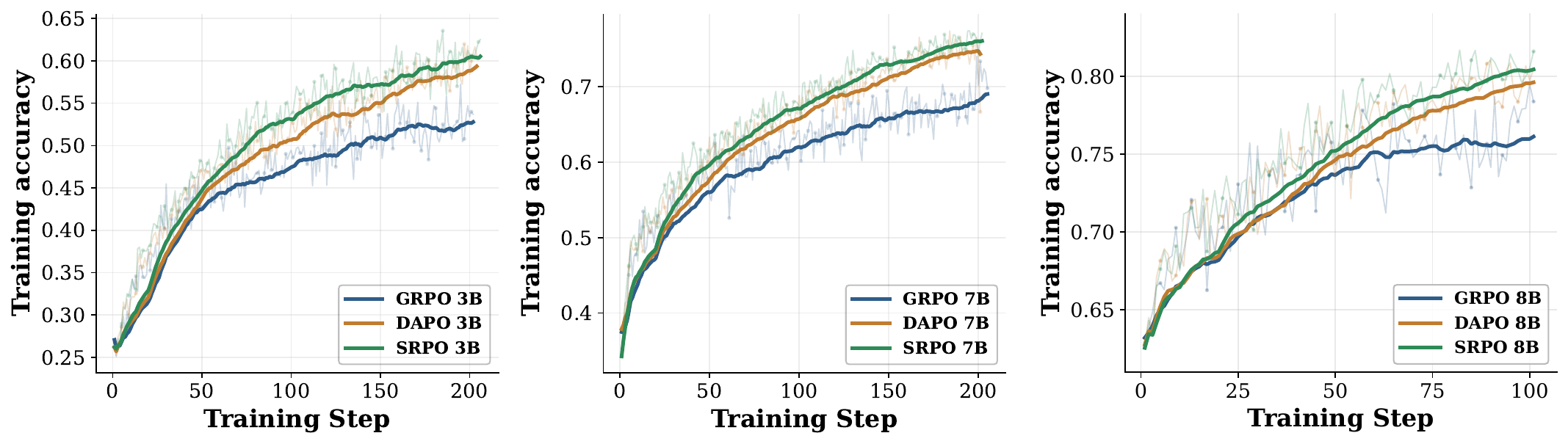}\vspace{-3mm}
    \caption{
Training accuracy of GRPO, DAPO, and SRPO on
Qwen2.5-VL-3B (left), Qwen2.5-VL-7B (middle),
and Qwen3-VL-8B (right).
Faint lines show raw values; solid lines show trailing
moving averages over 20 training steps.
}\vspace{-2mm}
    \label{fig:training_dynamics}
\end{figure}

Table~\ref{tab:main_results} compares SRPO with base models,
standard RL baselines, and other multimodal reasoning methods.
We report scores averaged over four repeated greedy evaluations,
with equal weight assigned to each of the nine benchmarks.
SRPO achieves the highest overall scores of 48.12, 54.96,
and 62.30 on Qwen2.5-VL-3B, Qwen2.5-VL-7B, and Qwen3-VL-8B,
respectively, improving over the corresponding base models
by 5.90, 5.33, and 4.86 percentage points.
Against the controlled RL baselines, SRPO outperforms GRPO
by 5.11, 4.15, and 4.28 points and DAPO by 4.30, 2.71,
and 3.14 points.
It also surpasses the strongest competing method within
each backbone group by 2.18, 1.37, and 1.84 points,
demonstrating consistent aggregate gains across all three
backbones.
Figure~\ref{fig:training_dynamics} further compares the training accuracy
of SRPO, GRPO, and DAPO across three backbones.
SRPO achieves higher smoothed training accuracy across all three
backbones, with a clear advantage at the end of training.
These trends complement the downstream evaluations,
supporting the effectiveness of role-aware token-level
credit assignment across model scales and backbone families.


\subsection{Ablation Studies}
\label{sec:ablation}

\definecolor{basegray}{RGB}{241,243,248}
\definecolor{fullblue}{RGB}{225,239,255}

\begin{table}[t]
\centering
\footnotesize
\setlength{\tabcolsep}{3pt}
\renewcommand{\arraystretch}{1.10}
\caption{
Ablations of components, credit-signal design, and uncertainty-penalty strength on Qwen2.5-VL.
Math, General, and Avg. denote unweighted averages over five math, four general, and all nine benchmarks, respectively.
$\Delta$ denotes the change from the corresponding base model.
}
\label{tab:ablation_summary}
\vspace{-2mm}

\begin{tabularx}{\columnwidth}{>{\raggedright\arraybackslash}Xrrrrrr}
\toprule
\textbf{Setting}
& \textbf{Math} & $\boldsymbol{\Delta}$
& \textbf{General} & $\boldsymbol{\Delta}$
& \textbf{Avg.} & $\boldsymbol{\Delta}$ \\
\midrule

\rowcolor{basegray}
\textbf{Qwen2.5-VL-3B}
& 36.96 & --
& 48.80 & --
& 42.22 & -- \\

\addlinespace[2pt]
\multicolumn{7}{l}{\textit{Component ablations}} \\
w/o perception credit
& 41.63 & $+4.67$
& 49.89 & $+1.09$
& 45.30 & $+3.08$ \\
w/o reasoning credit
& 41.10 & $+4.14$
& 48.75 & $-0.05$
& 44.50 & $+2.28$ \\
w/o unified modulation
& 40.03 & $+3.07$
& 46.98 & $-1.82$
& 43.12 & $+0.90$ \\

\addlinespace[2pt]
\multicolumn{7}{l}{\textit{Credit-signal design}} \\
w/ visual credit for reasoning
& 44.14 & $+7.18$
& 51.23 & $+2.43$
& 47.30 & $+5.08$ \\

\addlinespace[2pt]
\multicolumn{7}{l}{\textit{Uncertainty-penalty strength}} \\
$\eta=0$
& 40.55 & $+3.59$
& 48.58 & $-0.22$
& 44.12 & $+1.90$ \\
$\eta=0.1$
& 42.36 & $+5.40$
& 50.04 & $+1.24$
& 45.77 & $+3.55$ \\

\addlinespace[2pt]
\rowcolor{fullblue}
\textbf{SRPO ($\eta=0.03$)}
& \textbf{44.46} & $\mathbf{+7.50}$
& \textbf{52.71} & $\mathbf{+3.91}$
& \textbf{48.12} & $\mathbf{+5.90}$ \\

\midrule

\rowcolor{basegray}
\textbf{Qwen2.5-VL-7B}
& 44.78 & --
& 55.70 & --
& 49.63 & -- \\

\addlinespace[2pt]
\multicolumn{7}{l}{\textit{Component ablations}} \\
w/o perception credit
& 48.78 & $+4.00$
& 55.19 & $-0.51$
& 51.63 & $+2.00$ \\
w/o reasoning credit
& 48.25 & $+3.47$
& 54.08 & $-1.62$
& 50.84 & $+1.21$ \\
w/o unified modulation
& 46.74 & $+1.96$
& 53.30 & $-2.40$
& 49.65 & $+0.02$ \\

\addlinespace[2pt]
\multicolumn{7}{l}{\textit{Credit-signal design}} \\
w/ visual credit for reasoning
& 50.34 & $+5.56$
& 57.52 & $+1.82$
& 53.53 & $+3.90$ \\

\addlinespace[2pt]
\multicolumn{7}{l}{\textit{Uncertainty-penalty strength}} \\
$\eta=0$
& 47.68 & $+2.90$
& 54.47 & $-1.23$
& 50.70 & $+1.07$ \\
$\eta=0.1$
& 50.29 & $+5.51$
& 55.76 & $+0.06$
& 52.72 & $+3.09$ \\

\addlinespace[2pt]
\rowcolor{fullblue}
\textbf{SRPO ($\eta=0.03$)}
& \textbf{51.33} & $\mathbf{+6.55}$
& \textbf{59.49} & $\mathbf{+3.79}$
& \textbf{54.96} & $\mathbf{+5.33}$ \\

\bottomrule
\end{tabularx}

\vspace{-4mm}
\end{table}

Table~\ref{tab:ablation_summary} evaluates SRPO's components,
credit-signal design, and uncertainty-penalty strength.
We first remove perception credit, reasoning credit, or
unified modulation to assess the contributions of the two
role-specific signals and their shared trajectory-level
calibration.
Removing either credit signal degrades performance, with
a larger decline when reasoning credit is removed.
Removing unified modulation causes the largest degradation
among these component ablations, supporting the importance
of jointly calibrating the two signals.
We next examine whether a common visual-dependency signal
can replace the role-specific design.
Specifically, we retain the perception-credit computation
and replace the reasoning-credit contrast between contexts
with and without the perception segment with a contrast
between the original and masked images.
The same rollout tokens are rescored under both visual
conditions, with the generated perception context retained.
This variant tests whether direct image sensitivity provides
as effective a credit signal for reasoning as support from
the perception context.
It outperforms omitting reasoning credit but remains below
full SRPO in both task categories, supporting credit
assignment tailored to the distinct information dependencies
of perception and reasoning.
Finally, we vary the response-token uncertainty-penalty
coefficient $\eta \in \{0, 0.03, 0.1\}$ in the final
objective, leaving the credit scores and modulation weights
unchanged, to assess the penalty's contribution and strength.
Both nonzero settings outperform $\eta=0$, while
$\eta=0.03$ achieves the highest scores across both task
categories and model scales, supporting the inclusion of
the penalty at the selected strength.

%
%

\subsection{Further Analysis}
\label{sec:further_analysis}

\begin{figure}[t]
    \centering
    \includegraphics[width=0.8\textwidth]{
        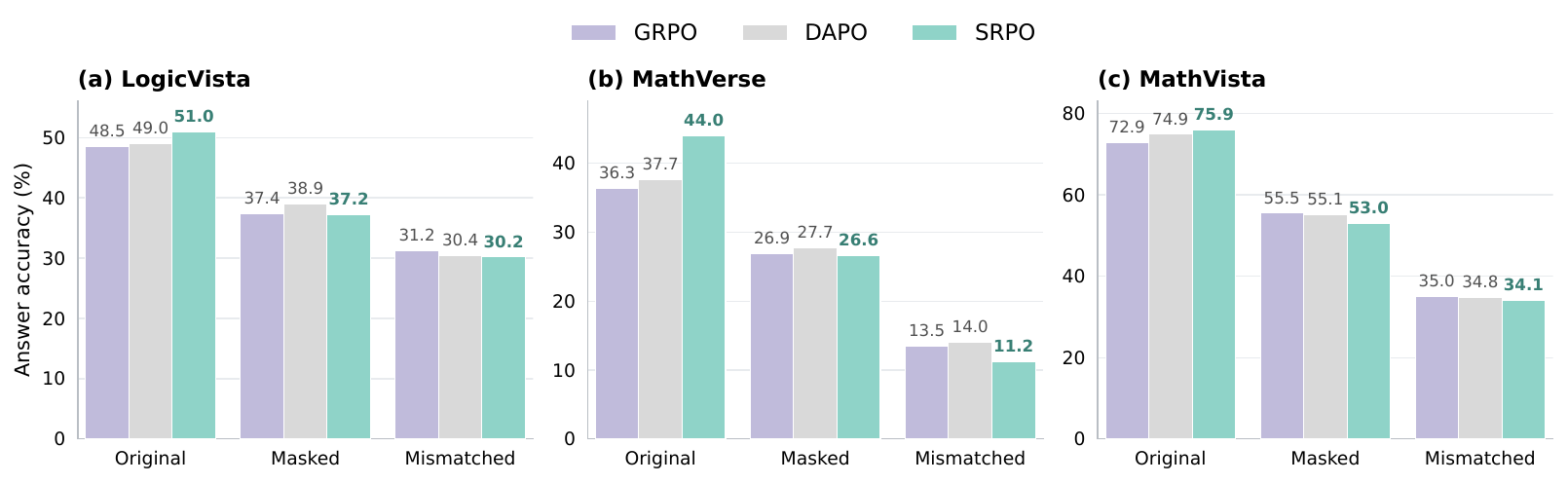
    }
    \vspace{-1mm}
    \caption{
    Answer accuracy of GRPO, DAPO, and SRPO with
    Qwen2.5-VL-7B under original, masked, and mismatched
    image conditions on LogicVista, MathVerse, and MathVista.
    }
    \label{fig:visual_interventions}
    \vspace{-4mm}
\end{figure}

\begin{figure}[t]
    \centering
    \includegraphics[width=0.7\columnwidth]{
        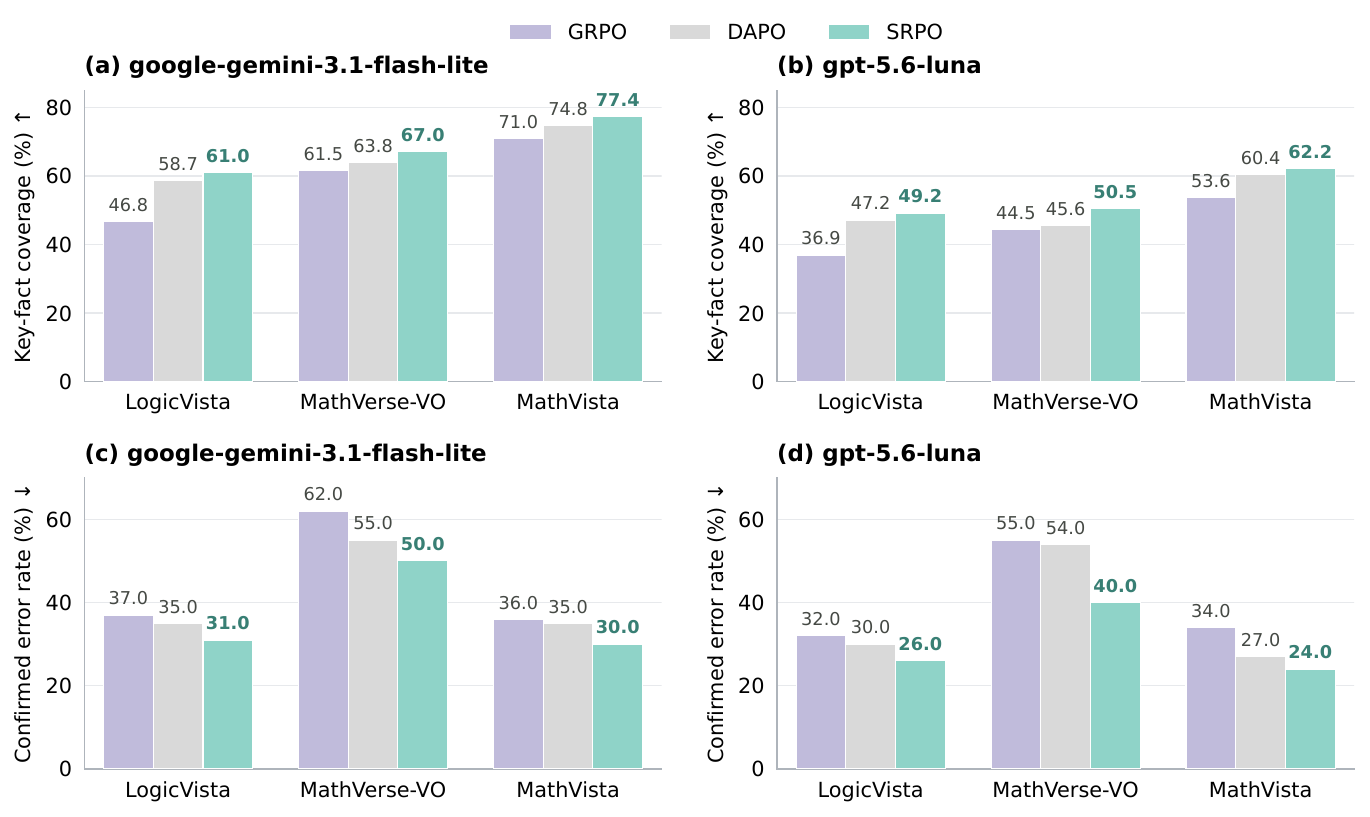
    }
    \vspace{-1mm}
    \caption{
    Perception-quality evaluation of GRPO, DAPO,
    and SRPO with Qwen2.5-VL-7B on the same 100 sampled
    questions per benchmark.
    The two columns show assessments by
    Gemini-3.1-flash-lite and GPT-5.6-luna.
    The top and bottom rows report reference key-fact coverage (higher is better) and the percentage of audited responses containing at least one judge-confirmed visual error (lower is better), respectively.
    }
    \label{fig:perception_audit}
    \vspace{-4mm}
\end{figure}

\paragraph{Sensitivity to visual interventions.}
Figure~\ref{fig:visual_interventions} compares final-answer
accuracy under original images, masked images that remove
visual information, and mismatched images that replace the
corresponding visual input.
We examine the trained Qwen2.5-VL-7B models on LogicVista,
MathVerse, and MathVista, regenerating
responses under each intervention.
Across the reported scores, SRPO achieves the highest
original-image accuracy on all three benchmarks but the
lowest accuracy under both masking and image mismatch.
Thus, its advantage with the original images does not
persist under the perturbed conditions.
However, the original-image scores are drawn from the main
evaluation, so attributing these differences specifically
to visual interventions requires matching the prompting,
decoding, and scoring protocols across conditions.
Moreover, greater performance degradation alone would not
establish more accurate perception, since it may also reflect
greater vulnerability to corrupted inputs.
We therefore complement this behavioral comparison with
a direct audit of the generated perception content.

\begin{figure}[t]
    \centering
    \includegraphics[width=0.4\columnwidth]{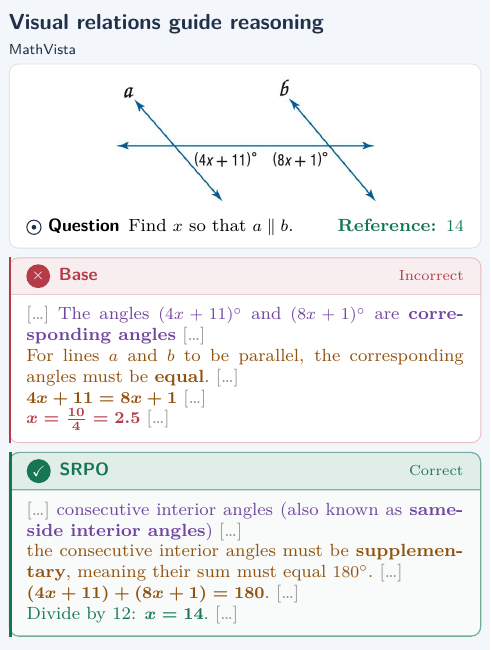}\vspace{-3mm}
    \caption{Case study illustrating how accurate visual relations support
correct reasoning.
The base model misidentifies same-side interior angles as
corresponding angles and equates their expressions,
whereas SRPO identifies the correct relation and applies
the supplementary-angle constraint.
    \textcolor[HTML]{7046A0}{Purple} highlights visual evidence and \textcolor[HTML]{92520E}{orange} highlights reasoning. \textcolor[HTML]{B53B48}{Red} and \textcolor[HTML]{167654}{green} indicate incorrect and correct answers, respectively.}\vspace{-5mm}
    \label{fig:case_study}
\end{figure}

\vspace{-2mm}
\paragraph{Perception quality.}
Figure~\ref{fig:perception_audit} evaluates the completeness
and correctness of intermediate perception content beyond
final-answer accuracy.
We compare the step-202 Qwen2.5-VL-7B checkpoints of GRPO,
DAPO, and SRPO on the same 100 sampled questions from each
of LogicVista, MathVerse, and MathVista, using a shared structured prompt, greedy decoding, and
a maximum output length of 2,048 tokens.
Two automated judges, google-gemini-3.1-flash-lite and
gpt-5.6-luna, separately assess the visual statements.
Key-fact coverage measures the fraction of reference visual
facts correctly extracted, averaged over eligible questions.
The confirmed perception-error rate measures the fraction
of all audited responses containing at least one identified
visual error.
Across all three benchmarks, both judges report higher
coverage and lower confirmed error rates for SRPO than
for GRPO and DAPO.
Relative to DAPO, SRPO improves coverage by
1.82--4.87 percentage points and reduces the confirmed
error rate by 3.00--14.00 points.
The accompanying coverage improvements indicate that
the lower confirmed error rates are not accompanied by
reduced coverage of the reference facts.
Together, these results support improved perception quality
under the automated auditing protocol, complementing the
answer-level evaluation.
They do not, however, establish that perception improvements
causally explain the downstream accuracy gains.

\subsection{Case Study}

Figure~\ref{fig:case_study} illustrates how visual relation recognition guides subsequent reasoning on MathVista. Both models read the angle expressions correctly, but Base identifies the angles as corresponding and equates them, obtaining $x=2.5$. SRPO identifies them as same-side interior angles and uses their supplementary relationship to obtain $x=14$. This behavior aligns with SRPO's role-aware training objective, which assigns perception credit based on visual dependency and reasoning credit based on consistency with the generated perception, while retaining final-answer supervision. In this example, the correctly identified angle relation provides the constraint used in the subsequent derivation. Additional paired examples are provided in Appendix~\ref{app:additional_case_study}.

\section{Conclusion}


In this paper, we present SRPO, a role-aware policy optimization method for multimodal RLVR.
SRPO structures responses into perception and reasoning, derives self-distilled token-level credit, and reweights GRPO advantages through bounded positive modulation. This reweighting preserves the advantage signs and retains the original verifiable reward without external reward models or separate teachers.
Experiments show improved training dynamics and downstream performance across multimodal reasoning benchmarks, with ablations supporting its role-aware design.

\newpage
\bibliography{wxg/main}

@String(ICLR = {Int. Conf. Learn. Represent.})

@String(AAAI = {AAAI})

@String(ICLR  = {ICLR})

@inproceedings{huang2026visionr1,
title={Vision-R1: Incentivizing Reasoning Capability in Multimodal Large Language Models},
author={Wenxuan Huang and Bohan Jia and Shaosheng Cao and Zheyu Ye and Fei zhao and Zhe Xu and Yao Hu and Shaohui Lin},
booktitle={The Fourteenth International Conference on Learning Representations},
year={2026}
}

@article{zhang2025r1,
  title={R1-VL: Learning to Reason with Multimodal Large Language Models via Step-wise Group Relative Policy Optimization},
  author={Zhang, Jingyi and Huang, Jiaxing and Yao, Huanjin and Liu, Shunyu and Zhang, Xikun and Lu, Shijian and Tao, Dacheng},
  journal={arXiv preprint arXiv:2503.12937},
  year={2025}
}

@inproceedings{yao2025r1,
  title={R1-ShareVL: Incentivizing Reasoning Capabilities of Multimodal Large Language Models via Share-GRPO},
  author={Yao, Huanjin and Yin, Qixiang and Zhang, Jingyi and Yang, Min and Wang, Yibo and Wu, Wenhao and Su, Fei and Shen, Li and Qiu, Minghui and Tao, Dacheng and others},
  booktitle={The Thirty-ninth Annual Conference on Neural Information Processing Systems},
  year={2025}
}

@inproceedings{
liu2026noisyrollout,
title={NoisyRollout: Reinforcing Visual Reasoning with Data Augmentation},
author={Xiangyan Liu and Jinjie Ni and Zijian Wu and Chao Du and Longxu Dou and Haonan Wang and Tianyu Pang and Michael Qizhe Shieh},
booktitle={The Thirty-ninth Annual Conference on Neural Information Processing Systems},
year={2026}
}

@inproceedings{
wang2026vlrethinker,
title={{VL}-Rethinker: Incentivizing Self-Reflection of Vision-Language Models with Reinforcement Learning},
author={Haozhe Wang and Chao Qu and Zuming Huang and Wei Chu and Fangzhen Lin and Wenhu Chen},
booktitle={The Thirty-ninth Annual Conference on Neural Information Processing Systems},
year={2026}
}

@article{wang2025skywork,
  title={Skywork r1v2: Multimodal hybrid reinforcement learning for reasoning},
  author={Wang, Peiyu and Wei, Yichen and Peng, Yi and Wang, Xiaokun and Qiu, Weijie and Shen, Wei and Xie, Tianyidan and Pei, Jiangbo and Zhang, Jianhao and Hao, Yunzhuo and others},
  journal={arXiv preprint arXiv:2504.16656},
  year={2025}
}

@article{xia2025visionary,
  title={Visionary-r1: Mitigating shortcuts in visual reasoning with reinforcement learning},
  author={Xia, Jiaer and Zang, Yuhang and Gao, Peng and Li, Sharon and Zhou, Kaiyang},
  journal={arXiv preprint arXiv:2505.14677},
  year={2025}
}

@article{xiao2025advancing,
  title={Perception-R1: Advancing Multimodal Reasoning Capabilities of MLLMs via Visual Perception Reward},
  author={Tong Xiao and Xin Xu and Zhenya Huang and Hongyu Gao and Quan Liu and Qi Liu and Enhong Chen},
  journal={arXiv preprint arXiv:2506.07218},
  year={2025},
}

@article{cao2025ground,
  title={Ground-r1: Incentivizing grounded visual reasoning via reinforcement learning},
  author={Cao, Meng and Zhao, Haoze and Zhang, Can and Chang, Xiaojun and Reid, Ian and Liang, Xiaodan},
  journal={arXiv preprint arXiv:2505.20272},
  year={2025}
}

@inproceedings{
sarch2026grounded,
title={Grounded Reinforcement Learning for Visual Reasoning},
author={Gabriel Herbert Sarch and Snigdha Saha and Naitik Khandelwal and Ayush Jain and Michael J. Tarr and Aviral Kumar and Katerina Fragkiadaki},
booktitle={The Thirty-ninth Annual Conference on Neural Information Processing Systems},
year={2026}
}

@inproceedings{
wang2026perceptionaware,
title={Perception-Aware Policy Optimization for Multimodal Reasoning},
author={Zhenhailong Wang and Xuehang Guo and Sofia Stoica and Haiyang Xu and Hongru WANG and Hyeonjeong Ha and Xiusi Chen and Yangyi Chen and Ming Yan and Fei Huang and Heng Ji},
booktitle={The Fourteenth International Conference on Learning Representations},
year={2026}
}

@inproceedings{
li2026visionsr,
title={Vision-{SR}1: Self-Rewarding Vision-Language Model via Reasoning Decomposition and Multi-Reward Policy Optimization},
author={Zongxia Li and Wenhao Yu and Chengsong Huang and Zhenwen Liang and Rui Liu and Fuxiao Liu and Jingxi Chen and Dian Yu and Jordan Lee Boyd-Graber and Haitao Mi and Dong Yu},
booktitle={The Fourteenth International Conference on Learning Representations},
year={2026}
}

@article{miao2026seeing,
  title={Seeing with You: Perception-Reasoning Coevolution for Multimodal Reasoning},
  author={Miao, Ziqi and Jia, Haonan and Li, Lijun and Qian, Chen and Xiong, Yuan and Yan, Wenting and Shao, Jing},
  journal={arXiv preprint arXiv:2603.28618},
  year={2026}
}

@article{wang2026visually,
  title={Visually-Guided Policy Optimization for Multimodal Reasoning},
  author={Wang, Zengbin and Xiong, Feng and Lin, Liang and Hu, Xuecai and Wang, Yong and Wang, Yanlin and Zhang, Man and Chu, Xiangxiang},
  journal={arXiv preprint arXiv:2604.09349},
  year={2026}
}

@inproceedings{
guo2026segment,
title={Segment Policy Optimization: Effective Segment-Level Credit Assignment in {RL} for Large Language Models},
author={Yiran Guo and Lijie Xu and Jie Liu and Ye Dan and Shuang Qiu},
booktitle={The Thirty-ninth Annual Conference on Neural Information Processing Systems},
year={2026}
}

@article{tran2025exploiting,
  title={Exploiting tree structure for credit assignment in rl training of llms},
  author={Tran, Hieu and Yao, Zonghai and Yu, Hong},
  journal={arXiv preprint arXiv:2509.18314},
  year={2025}
}

@inproceedings{
wang2026beyond,
title={Beyond the 80/20 Rule: High-Entropy Minority Tokens Drive Effective Reinforcement Learning for {LLM} Reasoning},
author={Shenzhi Wang and Le Yu and Chang Gao and Chujie Zheng and Shixuan Liu and Rui Lu and Kai Dang and Xiong-Hui Chen and Jianxin Yang and Zhenru Zhang and Yuqiong Liu and An Yang and Andrew Zhao and Yang Yue and Shiji Song and Bowen Yu and Gao Huang and Junyang Lin},
booktitle={The Thirty-ninth Annual Conference on Neural Information Processing Systems},
year={2026}
}

@article{lu2026bridging,
  title={Bridging Perception and Reasoning: Token Reweighting for RLVR in Multimodal LLMs},
  author={Lu, Jinda and Wu, Junkang and Li, Jinghan and Huang, Kexin and Yang, Shuo and Wang, Guoyin and Wu, Jiancan and Wang, Xiang and He, Xiangnan},
  journal={arXiv preprint arXiv:2603.25077},
  year={2026}
}

@article{ye2026not,
  title={Not All Tokens See Equally: Perception-Grounded Policy Optimization for Large Vision-Language Models},
  author={Ye, Zekai and Li, Qiming and Feng, Xiaocheng and Chen, Ruihan and Li, Ziming and Ren, Haoyu and Chen, Kun and Tu, Dandan and Qin, Bing},
  journal={arXiv preprint arXiv:2604.01840},
  year={2026}
}

@article{ding2025vtperception,
  title={VTPerception-R1: Enhancing Multimodal Reasoning via Explicit Visual and Textual Perceptual Grounding},
  author={Ding, Yizhuo and Chen, Mingkang and Feng, Zhibang and Xiao, Tong and Qu, Wanying and Shao, Wenqi and Fu, Yanwei},
  journal={arXiv preprint arXiv:2509.24776},
  year={2025}
}

@inproceedings{
huang2026spotlight,
title={Spotlight on Token Perception for Multimodal Reinforcement Learning},
author={Siyuan Huang and Xiaoye Qu and Yafu Li and Yun Luo and Zefeng He and Daizong Liu and Yu Cheng},
booktitle={The Fourteenth International Conference on Learning Representations},
year={2026}
}

@article{chen2025perception,
  title={Perception before reasoning: Two-stage reinforcement learning for visual reasoning in vision-language models},
  author={Chen, Yan and Li, Long and Xi, Teng and Zeng, Long and Wang, Jingdong},
  journal={arXiv preprint arXiv:2509.13031},
  year={2025}
}

@inproceedings{zha2026vision,
  title={Vision-G1: Towards General Reasoning Vision-Language Models via Reinforcement Learning},
  author={Zha, Yuheng and Zhou, Kun and Wu, Yujia and Wang, Yushu and Feng, Jie and Xu, Zhi and Hao, Shibo and Liu, Zhengzhong and Xing, Eric P and Hu, Zhiting},
  booktitle={Proceedings of the AAAI Conference on Artificial Intelligence},
  year={2026}
}

@inproceedings{
yu2026perceptionr,
title={Perception-R1: Pioneering Perception Policy with Reinforcement Learning},
author={En Yu and Kangheng Lin and Liang Zhao and jisheng yin and Yana Wei and Yuang Peng and Haoran Wei and Jianjian Sun and Chunrui Han and Zheng Ge and Xiangyu Zhang and Daxin Jiang and Jingyu Wang and Wenbing Tao},
booktitle={The Thirty-ninth Annual Conference on Neural Information Processing Systems},
year={2026}
}

@article{xu2025kdrl,
  title={Kdrl: Post-training reasoning llms via unified knowledge distillation and reinforcement learning},
  author={Xu, Hongling and Zhu, Qi and Deng, Heyuan and Li, Jinpeng and Hou, Lu and Wang, Yasheng and Shang, Lifeng and Xu, Ruifeng and Mi, Fei},
  journal={arXiv preprint arXiv:2506.02208},
  year={2025}
}

@article{zhang2026reinforcement,
  title={Reinforcement-aware knowledge distillation for LLM reasoning},
  author={Zhang, Zhaoyang and Jiang, Shuli and Shen, Yantao and Zhang, Yuting and Ram, Dhananjay and Yang, Shuo and Tu, Zhuowen and Xia, Wei and Soatto, Stefano},
  journal={arXiv preprint arXiv:2602.22495},
  year={2026}
}

@inproceedings{
su2026pixel,
title={Pixel Reasoner: Incentivizing Pixel Space Reasoning via Curiosity-Driven Reinforcement Learning},
author={Alex Su and Haozhe Wang and Weiming Ren and Fangzhen Lin and Wenhu Chen},
booktitle={The Thirty-ninth Annual Conference on Neural Information Processing Systems},
year={2026},
}

@inproceedings{
sun2026regionreasoner,
title={RegionReasoner: Region-Grounded Multi-Round Visual Reasoning},
author={Wenfang Sun and Hao Chen and Yingjun Du and Yefeng Zheng and Cees G. M. Snoek},
booktitle={The Fourteenth International Conference on Learning Representations},
year={2026}
}

@article{yang2026self,
  title={Self-Distilled RLVR},
  author={Yang, Chenxu and Qin, Chuanyu and Si, Qingyi and Chen, Minghui and Gu, Naibin and Yao, Dingyu and Lin, Zheng and Wang, Weiping and Wang, Jiaqi and Duan, Nan},
  journal={arXiv preprint arXiv:2604.03128},
  year={2026}
}

@inproceedings{
liu2026visionreasoner,
title={VisionReasoner: Unified Reasoning-Integrated Visual Perception via Reinforcement Learning},
author={Yuqi Liu and Tianyuan Qu and Zhisheng Zhong and Bohao PENG and Shu Liu and Bei Yu and Jiaya Jia},
booktitle={The Fourteenth International Conference on Learning Representations},
year={2026}
}

@article{he2025visplay,
  title={Visplay: Self-evolving vision-language models from images},
  author={He, Yicheng and Huang, Chengsong and Li, Zongxia and Huang, Jiaxin and Yang, Yonghui},
  journal={arXiv preprint arXiv:2511.15661},
  year={2025}
}

@article{Qwen2.5-VL,
  title={Qwen2.5-VL Technical Report},
  author={Bai, Shuai and Chen, Keqin and Liu, Xuejing and Wang, Jialin and Ge, Wenbin and Song, Sibo and Dang, Kai and Wang, Peng and Wang, Shijie and Tang, Jun and Zhong, Humen and Zhu, Yuanzhi and Yang, Mingkun and Li, Zhaohai and Wan, Jianqiang and Wang, Pengfei and Ding, Wei and Fu, Zheren and Xu, Yiheng and Ye, Jiabo and Zhang, Xi and Xie, Tianbao and Cheng, Zesen and Zhang, Hang and Yang, Zhibo and Xu, Haiyang and Lin, Junyang},
  journal={arXiv preprint arXiv:2502.13923},
  year={2025}
}

@misc{deepseek-math,
  author = {Zhihong Shao and Peiyi Wang and Qihao Zhu and Runxin Xu and Junxiao Song and Mingchuan Zhang and Y.K. Li and Y. Wu and Daya Guo},
  title = {DeepSeekMath: Pushing the Limits of Mathematical Reasoning in Open Language Models},
  journal = {CoRR},
  volume = {abs/2402.03300},
  year = {2024}
}

@inproceedings{
yu2026dapo,
title={{DAPO}: An Open-Source {LLM} Reinforcement Learning System at Scale},
author={Qiying Yu and Zheng Zhang and Ruofei Zhu and Yufeng Yuan and Xiaochen Zuo and YuYue and Weinan Dai and Tiantian Fan and Gaohong Liu and Juncai Liu and LingJun Liu and Xin Liu and Haibin Lin and Zhiqi Lin and Bole Ma and Guangming Sheng and Yuxuan Tong and Chi Zhang and Mofan Zhang and Ru Zhang and Wang Zhang and Hang Zhu and Jinhua Zhu and Jiaze Chen and Jiangjie Chen and Chengyi Wang and Hongli Yu and Yuxuan Song and Xiangpeng Wei and Hao Zhou and Jingjing Liu and Wei-Ying Ma and Ya-Qin Zhang and Lin Yan and Yonghui Wu and Mingxuan Wang},
booktitle={The Thirty-ninth Annual Conference on Neural Information Processing Systems},
year={2026}
}

@article{li2025vision,
  title={Vision Matters: Simple Visual Perturbations Can Boost Multimodal Math Reasoning},
  author={Li, Yuting and Wei, Lai and Zheng, Kaipeng and Huang, Jingyuan and Kong, Linghe and Sun, Lichao and Huang, Weiran},
  journal={arXiv preprint arXiv:2506.09736},
  year={2025}
}

@misc{zheng2025easyr1,
  title        = {EasyR1: An Efficient, Scalable, Multi-Modality RL Training Framework},
  author       = {Yaowei Zheng and Junting Lu and Shenzhi Wang and Zhangchi Feng and Dongdong Kuang and Yuwen Xiong and Richong Zhang},
  year         = {2025}
}

@article{loshchilov2017decoupled,
  title={Decoupled weight decay regularization},
  author={Loshchilov, Ilya and Hutter, Frank},
  journal={arXiv preprint arXiv:1711.05101},
  year={2017}
}

@ARTICLE{2024arXiv240902813Y,
       author = {{Yue}, Xiang and {Zheng}, Tianyu and {Ni}, Yuansheng and {Wang}, Yubo and {Zhang}, Kai and {Tong}, Shengbang and {Sun}, Yuxuan and {Yu}, Botao and {Zhang}, Ge and {Sun}, Huan and {Su}, Yu and {Chen}, Wenhu and {Neubig}, Graham},
        title = "{MMMU-Pro: A More Robust Multi-discipline Multimodal Understanding Benchmark}",
         year = 2024,
          eid = {arXiv:2409.02813},
archivePrefix = {arXiv},
       eprint = {2409.02813}
}

@article{qiao2024we,
  title={We-Math: Does Your Large Multimodal Model Achieve Human-like Mathematical Reasoning?},
  author={Qiao, Runqi and Tan, Qiuna and Dong, Guanting and Wu, Minhui and Sun, Chong and Song, Xiaoshuai and GongQue, Zhuoma and Lei, Shanglin and Wei, Zhe and Zhang, Miaoxuan and others},
  journal={arXiv preprint arXiv:2407.01284},
  year={2024}
}

@misc{xiao2024logicvistamultimodalllmlogical,
      title={LogicVista: Multimodal LLM Logical Reasoning Benchmark in Visual Contexts}, 
      author={Yijia Xiao and Edward Sun and Tianyu Liu and Wei Wang},
      year={2024},
      eprint={2407.04973},
      archivePrefix={arXiv}
    
}

@inproceedings{
wang2024measuring,
title={Measuring Multimodal Mathematical Reasoning with MATH-Vision Dataset},
author={Ke Wang and Junting Pan and Weikang Shi and Zimu Lu and Houxing Ren and Aojun Zhou and Mingjie Zhan and Hongsheng Li},
booktitle={The Thirty-eight Conference on Neural Information Processing Systems Datasets and Benchmarks Track},
year={2024}
}

@article{zhang2024mathverse,
  title={MathVerse: Does Your Multi-modal LLM Truly See the Diagrams in Visual Math Problems?},
  author={Zhang, Renrui and Jiang, Dongzhi and Zhang, Yichi and Lin, Haokun and Guo, Ziyu and Qiu, Pengshuo and Zhou, Aojun and Lu, Pan and Chang, Kai-Wei and Gao, Peng and others},
  journal={arXiv preprint arXiv:2403.14624},
  year={2024}
}

@inproceedings{duan2024vlmevalkit,
  title={Vlmevalkit: An open-source toolkit for evaluating large multi-modality models},
  author={Duan, Haodong and Yang, Junming and Qiao, Yuxuan and Fang, Xinyu and Chen, Lin and Liu, Yuan and Dong, Xiaoyi and Zang, Yuhang and Zhang, Pan and Wang, Jiaqi and others},
  booktitle={Proceedings of the 32nd ACM International Conference on Multimedia},
  pages={11198--11201},
  year={2024}
}

@inproceedings{
zou2025dynamath,
title={DynaMath: A Dynamic Visual Benchmark for Evaluating Mathematical Reasoning Robustness of Vision Language Models},
author={Chengke Zou and Xingang Guo and Rui Yang and Junyu Zhang and Bin Hu and Huan Zhang},
booktitle={The Thirteenth International Conference on Learning Representations},
year={2025}
}

@inproceedings{yu2024mm,
  title={Mm-vet: Evaluating large multimodal models for integrated capabilities},
  author={Yu, Weihao and Yang, Zhengyuan and Li, Linjie and Wang, Jianfeng and Lin, Kevin and Liu, Zicheng and Wang, Xinchao and Wang, Lijuan},
  booktitle={International conference on machine learning},
  year={2024},
  organization={PMLR}
}

@article{guo2025deepseek,
  title={Deepseek-r1: Incentivizing reasoning capability in llms via reinforcement learning},
  author={Guo, Daya and Yang, Dejian and Zhang, Haowei and Song, Junxiao and Wang, Peiyi and Zhu, Qihao and Xu, Runxin and Zhang, Ruoyu and Ma, Shirong and Bi, Xiao and others},
  journal={arXiv preprint arXiv:2501.12948},
  year={2025}
}

@inproceedings{
ghosh2025visual,
title={Visual Description Grounding Reduces Hallucinations and Boosts Reasoning in {LVLM}s},
author={Sreyan Ghosh and Chandra Kiran Reddy Evuru and Sonal Kumar and Utkarsh Tyagi and Oriol Nieto and Zeyu Jin and Dinesh Manocha},
booktitle={The Thirteenth International Conference on Learning Representations},
year={2025}
}

@inproceedings{
fang2026grounding,
title={Grounding Language with Vision: A Conditional Mutual Information Calibrated Decoding Strategy for Reducing Hallucinations in {LVLM}s},
author={Hao Fang and Changle Zhou and Jiawei Kong and Kuofeng Gao and Bin Chen and Tao Liang and Guojun Ma and Shu-Tao Xia},
booktitle={The Thirty-ninth Annual Conference on Neural Information Processing Systems},
year={2026}
}

@inproceedings{sharma2026see,
  title={See, think, learn: A self-taught multimodal reasoner},
  author={Sharma, Sourabh and Gupta, Sonam and Sadbhawna, Sadbhawna},
  booktitle={Proceedings of the IEEE/CVF Winter Conference on Applications of Computer Vision},
  year={2026}
}

@inproceedings{zhou2025proreason,
  title={PROREASON: Multi-Modal Proactive Reasoning with Decoupled Eyesight and Wisdom},
  author={Zhou, Jingqi and Wang, Sheng and Dong, Jingwei and Liu, Kai and Li, Lei and Gao, Jiahui and Jiang, Jiyue and Kong, Lingpeng and Wu, Chuan},
  booktitle={Proceedings of the 2025 Conference on Empirical Methods in Natural Language Processing},
  year={2025}
}

@inproceedings{
li2024naturalbench,
title={NaturalBench: Evaluating Vision-Language Models on Natural Adversarial Samples},
author={Baiqi Li and Zhiqiu Lin and Wenxuan Peng and Jean de Dieu Nyandwi and Daniel Jiang and Zixian Ma and Simran Khanuja and Ranjay Krishna and Graham Neubig and Deva Ramanan},
booktitle={The Thirty-eight Conference on Neural Information Processing Systems Datasets and Benchmarks Track},
year={2024}
}

@ARTICLE{2024arXiv240305530G,
  author        = {{Gemini Team} and Georgiev, Petko and Lei, Ving Ian and Burnell, Ryan and Bai, Libin and others},
  title         = "{Gemini 1.5: Unlocking multimodal understanding across millions of tokens of context}",
  journal       = {arXiv e-prints},
  year          = {2024},
  eid           = {arXiv:2403.05530},
  archivePrefix = {arXiv},
  eprint        = {2403.05530},
}

@article{achiam2023gpt,
  title={Gpt-4 technical report},
  author={Achiam, Josh and Adler, Steven and Agarwal, Sandhini and Ahmad, Lama and Akkaya, Ilge and Aleman, Florencia Leoni and Almeida, Diogo and Altenschmidt, Janko and Altman, Sam and Anadkat, Shyamal and others},
  journal={arXiv preprint arXiv:2303.08774},
  year={2023}
}

@article{zhu2025internvl3,
  title={Internvl3: Exploring advanced training and test-time recipes for open-source multimodal models},
  author={Zhu, Jinguo and Wang, Weiyun and Chen, Zhe and Liu, Zhaoyang and Ye, Shenglong and Gu, Lixin and Tian, Hao and Duan, Yuchen and Su, Weijie and Shao, Jie and others},
  journal={arXiv preprint arXiv:2504.10479},
  year={2025}
}

@inproceedings{lu2024mathvista,
  title={MathVista: Evaluating Mathematical Reasoning of Foundation Models in Visual Contexts},
  author={Lu, Pan and Bansal, Hritik and Xia, Tony and Liu, Jiacheng and Li, Chunyuan and Hajishirzi, Hannaneh and Cheng, Hao and Chang, Kai-Wei and Galley, Michel and Gao, Jianfeng},
  booktitle={International Conference on Learning Representations (ICLR)},
  year={2024}
}

@inproceedings{
lightman2023lets,
title={Let's Verify Step by Step},
author={Hunter Lightman and Vineet Kosaraju and Yuri Burda and Harrison Edwards and Bowen Baker and Teddy Lee and Jan Leike and John Schulman and Ilya Sutskever and Karl Cobbe},
booktitle={The Twelfth International Conference on Learning Representations},
year={2024}
}

@misc{wang2025visualprm,
      title={VisualPRM: An Effective Process Reward Model for Multimodal Reasoning}, 
      author={Weiyun Wang and Zhangwei Gao and Lianjie Chen and Zhe Chen and Jinguo Zhu and Xiangyu Zhao and Yangzhou Liu and Yue Cao and Shenglong Ye and Xizhou Zhu and Lewei Lu and Haodong Duan and Yu Qiao and Jifeng Dai and Wenhai Wang},
      year={2025},
      eprint={2503.10291},
      archivePrefix={arXiv},
      primaryClass={cs.CV}
}

@misc{wang2024mathshepherd,
      title={Math-Shepherd: Verify and Reinforce LLMs Step-by-step without Human Annotations}, 
      author={Peiyi Wang and Lei Li and Zhihong Shao and R. X. Xu and Damai Dai and Yifei Li and Deli Chen and Y. Wu and Zhifang Sui},
      year={2024},
      eprint={2312.08935},
      archivePrefix={arXiv},
      primaryClass={cs.AI}
}

@article{Qwen3-VL,
      title={Qwen3-VL Technical Report}, 
      author={Shuai Bai and Yuxuan Cai and Ruizhe Chen and Keqin Chen and Xionghui Chen and Zesen Cheng and Lianghao Deng and Wei Ding and Chang Gao and Chunjiang Ge and Wenbin Ge and Zhifang Guo and Qidong Huang and Jie Huang and Fei Huang and Binyuan Hui and Shutong Jiang and Zhaohai Li and Mingsheng Li and Mei Li and Kaixin Li and Zicheng Lin and Junyang Lin and Xuejing Liu and Jiawei Liu and Chenglong Liu and Yang Liu and Dayiheng Liu and Shixuan Liu and Dunjie Lu and Ruilin Luo and Chenxu Lv and Rui Men and Lingchen Meng and Xuancheng Ren and Xingzhang Ren and Sibo Song and Yuchong Sun and Jun Tang and Jianhong Tu and Jianqiang Wan and Peng Wang and Pengfei Wang and Qiuyue Wang and Yuxuan Wang and Tianbao Xie and Yiheng Xu and Haiyang Xu and Jin Xu and Zhibo Yang and Mingkun Yang and Jianxin Yang and An Yang and Bowen Yu and Fei Zhang and Hang Zhang and Xi Zhang and Bo Zheng and Humen Zhong and Jingren Zhou and Fan Zhou and Jing Zhou and Yuanzhi Zhu and Ke Zhu},
	  journal={arXiv preprint arXiv:2511.21631},
      year={2025}
}

\newpage
\appendix

\begin{center}
    {\Large\bfseries Appendix}
\end{center}



\section{Training Procedure}
\label{app:training_procedure}

For clarity and reproducibility, we provide a step-by-step description of the SRPO training procedure in Algorithm~\ref{alg:srca_training}.
The algorithm elaborates on the role-aware credit assignment mechanism introduced in Section~\ref{sec:srca}.
Each training iteration consists of four main stages.
First, the behavior policy generates a group of on-policy responses for each image--question pair under the structured perception--reasoning template.
Second, a verifier assigns outcome-level rewards based on format compliance and final-answer correctness, from which group-normalized GRPO advantages are computed.
Third, SRPO parses each response into perception and reasoning spans and performs stop-gradient rescoring with the behavior policy under controlled context changes: perception tokens are rescored under the original and masked images to estimate visual dependency, while reasoning tokens are rescored with and without the generated perception to estimate perception-supported grounding consistency.
Finally, the role-specific scores are centered within each trajectory, converted into bounded positive token-level modulation weights, and used to form token-level advantages for the PPO-style policy update.
The response-token entropy surrogate is added only in the final optimization objective and does not affect the stop-gradient credit scores or modulation weights.

\section{Implementation and Evaluation Details}
\label{app:implementation_details}

\subsection{Implementation Details}
\label{app:training_details}

\paragraph{Overall setup.}
Our implementation uses EasyR1~\citep{zheng2025easyr1} with Qwen2.5-VL-3B-Instruct, Qwen2.5-VL-7B-Instruct, and Qwen3-VL-8B-Instruct as backbones.
SRPO and the controlled GRPO and DAPO baselines are trained on 8 NVIDIA H100 GPUs using AdamW with a learning rate of $1\times10^{-6}$.
They share the training data, structured response template, base reward, rollout budget, and optimizer settings; their KL regularization, clipping, and filtering configurations are specified below.
The profiled SRPO run on eight NVIDIA H20 GPUs takes approximately 17.69 hours (141.52 GPU-hours).

\begin{algorithm}[t]
\caption{SRPO: Structured Role-aware Policy Optimization}
\label{alg:srca_training}
\begin{algorithmic}[1]
\Require Policy $\pi_\theta$, dataset $\mathcal{D}$, group size $G$, masking operator $\mathcal{M}$.
\Ensure Updated policy $\pi_\theta$.

\For{each training iteration}
    \State Set the behavior policy $\pi_{\theta_{\mathrm{old}}}\leftarrow \pi_\theta$.
    \State Sample a batch of training instances $(I,q,a)\sim\mathcal{D}$.
    \Statex \hfill $\triangleright$ \textbf{Phase 1: On-policy Rollout Generation}
    \For{each $(I,q,a)$}
        \State Sample $G$ structured rollouts
        $\{y_i=(y_i^{\mathrm{perc}},y_i^{\mathrm{reas}})\}_{i=1}^{G}\sim \pi_{\theta_{\mathrm{old}}}(\cdot\mid I,q)$.
        \State Compute verifier rewards $R_i$ and group-relative advantages $\hat A_i$.
    \EndFor
    \Statex \hfill $\triangleright$ \textbf{Phase 2: Role-specific Stop-gradient Rescoring}
    \For{each trajectory $y_i$}
        \State Parse perception positions $\mathcal{P}_i$, reasoning positions $\mathcal{R}_i$, and valid positions $\mathcal{V}(y_i)$.
        \State Construct a masked image $I^{\mathrm{mask}}=\mathcal{M}(I)$.
        \For{each $t\in\mathcal{V}(y_i)$}
            \If{$t\in\mathcal{P}_i$}
                \State Compute $\delta_{i,t}^{\mathrm{perc}}$ by Eq.~\eqref{eq:perc_delta} using $\pi_{\theta_{\mathrm{old}}}$.
                \State Compute $r_{i,t}^{\mathrm{perc}}$ by Eq.~\eqref{eq:perc_prop_weight}.
                \State Set $r_{i,t}\leftarrow r_{i,t}^{\mathrm{perc}}$.
            \ElsIf{$t\in\mathcal{R}_i$}
                \State Compute $\delta_{i,t}^{\mathrm{reas}}$ by Eq.~\eqref{eq:reas_delta} using $\pi_{\theta_{\mathrm{old}}}$.
                \State Compute $r_{i,t}^{\mathrm{reas}}$ by Eq.~\eqref{eq:reas_modulation}.
                \State Set $r_{i,t}\leftarrow r_{i,t}^{\mathrm{reas}}$.
            \EndIf
        \EndFor
    \EndFor
    \Statex \hfill $\triangleright$ \textbf{Phase 3: Unified Trajectory-level Modulation}
    
    \For{each trajectory $y_i$}
        \State Compute the response-level baseline
        $
        \bar r_i
        =
        |\mathcal{V}(y_i)|^{-1}
        \sum_{t\in\mathcal{V}(y_i)} r_{i,t}$.
        
        \For{each $t\in\mathcal{V}(y_i)$}
            \State Compute $w_{i,t}$ by Eq.~\eqref{eq:simple_mod}.
            \State Form the token-level advantage
            $
            \tilde A_{i,t}=\hat A_i w_{i,t}$.
        \EndFor
    \EndFor
    \Statex \hfill $\triangleright$ \textbf{Phase 4: Policy Optimization}
    
    \State Update $\theta$ by maximizing the total objective in Eq.~\eqref{eq:srca_total_objective}.
\EndFor
\end{algorithmic}
\end{algorithm}

\paragraph{RL training configuration.}
We train Qwen2.5-VL models for two epochs and Qwen3-VL-8B for one epoch on ViRL39K~\citep{wang2026vlrethinker}, using the final checkpoint from each run.
The rollout batch size is 384, the actor global batch size is 128, and the validation batch size is 512.
For each prompt, we sample 8 responses at temperature 1.0 and top-$p$ 0.99.
The maximum response length is 2,048 tokens for Qwen2.5-VL-3B/7B and 4,096 for Qwen3-VL-8B.
GRPO, DAPO, and SRPO use the same base reward, $R_i=0.95R_i^{\mathrm{acc}}+0.05R_i^{\mathrm{fmt}}$.
The accuracy reward is 1 if the extracted final answer matches the ground truth under the equivalence checker, and 0 otherwise.
GRPO uses a reference-policy KL penalty with coefficient 0.01 and disables online filtering.
Its lower and upper clipping parameters are 0.20/0.30 for Qwen2.5-VL and 0.20/0.20 for Qwen3-VL.
DAPO and SRPO use clipping parameters of 0.20/0.28, enable online filtering, and omit the reference-policy KL penalty on all three backbones.
The visual-credit-for-reasoning ablation retains the SRPO optimization configuration.

\paragraph{Self-distilled credit assignment.}

All credit scores are computed by rescoring responses
sampled under the original image and structured prompt.
We keep the target tokens fixed and evaluate their
log-probabilities under modified contexts using the
behavior policy with stop-gradient, without generating
new responses.

For perception-token credit assignment, SRPO constructs
a weak visual branch through random patch blackening:
the image is divided into non-overlapping $14\times14$
patches, each independently masked with probability
0.5~\citep{wang2026perceptionaware}.
We compute perception credit from the token
log-probability contrast between the original and masked
images, retaining the same question and response prefix.
The perception scores are then jointly calibrated with
the reasoning scores through unified trajectory-level
modulation.

For reasoning-token credit assignment, SRPO compares each reasoning token under two textual conditioning contexts.
The first is the full rollout context, which includes the image, question, generated perception, and previous reasoning tokens.
The second is a perception-ablated context, which removes the generated perception block and its corresponding instruction while retaining the image, question content, and reasoning prefix.
The resulting contrast measures the change in token likelihood between the full and perception-ablated contexts.
The perception-ablated scoring branch uses the template \texttt{qi\_only\_reasoning\_answer.jinja}.
The resulting residual is modulated by the sign of the trajectory-level advantage and mapped to a positive token-level factor through an exponential transformation.
In the unified trajectory-level modulation, we set $\lambda_{\mathrm{mod}}=0.5$ and clip the final token weights to $[0.8,1.2]$.

\paragraph{Additional actor regularization.}
To improve the stability of actor optimization, we include a lightweight sampled-token negative-log-probability regularizer in the actor loss:
\begin{equation}
\label{eq:actor_nll_regularizer}
\mathcal{L}_{\mathrm{reg}}
=
\eta\,
\mathbb{E}_{(s_t,a_t)}
\left[
-\log \pi_{\theta}(a_t\mid s_t)
\right],
\end{equation}
where the expectation is taken over valid response tokens sampled during rollout, and $\eta=0.03$.
This term regularizes the current policy along the sampled response trajectory and is applied only during actor optimization.
It is not used to compute the self-distilled token scores or the modulation weights.
We note that this term is a sampled-token entropy surrogate rather than a full-vocabulary Shannon entropy.

\subsection{Evaluation Details}
\label{app:evaluation_details}

\paragraph{Evaluation framework.}
We use VLMEvalKit~\citep{duan2024vlmevalkit} for benchmark
loading, input construction, answer extraction, and scoring,
with method-specific prompts as described below.
All main-table results are obtained through our own
evaluation runs rather than copied from prior papers.
We evaluate publicly released checkpoints for external
baselines, except for Qwen3-VL-8B PAPO-G and PAPO-D,
which we reproduce using their original training prompts.
PRCO follows its published inference procedure and
method-specific settings.

\begin{table}[t]
  \centering
  \caption{Vision-language benchmarks evaluated in VLMEvalKit.}
  \label{tab:math-benchmarks-vlmevalkit}
  \small
  \setlength{\tabcolsep}{6pt}
  \resizebox{0.7\linewidth}{!}{%
    \begin{tabular}{@{}l r l@{}}
      \toprule
      \textbf{Benchmark} & \textbf{\#Samples} & \textbf{Metric} \\
      \midrule
      MathVista\_MINI              & 1000 & Overall \\
      MathVision\_MINI             & 304  & Overall \\
      WeMath                       & 1740 & Score (Strict) \\
      MathVerse\_MINI\_Vision\_Only & 788  & Overall \\
      DynaMath                     & 5010 & Overall (Average) \\
      MMMU\_Pro\_V                 & 1730 & Overall \\
      MMVet                        & 218  & Overall \\
      LogicVista                   & 447  & Overall \\
      NaturalBench                 & 7600 (1900 groups) & Overall \\
      \bottomrule
    \end{tabular}%
  }
\end{table}

\paragraph{Benchmarks.}

Our evaluation covers nine vision-language benchmarks using VLMEvalKit~\citep{duan2024vlmevalkit}, spanning math-centric reasoning, multidisciplinary multimodal understanding, logical reasoning, integrated visual reasoning, and robustness evaluation. 
Table~\ref{tab:math-benchmarks-vlmevalkit} summarizes the dataset sizes and reported metrics, and we briefly describe the evaluation focus of each benchmark below.

\begin{itemize}[leftmargin=*]
    \item \textbf{MathVista}~\citep{lu2024mathvista} is a comprehensive benchmark for visual mathematical reasoning. 
It covers diverse visual contexts such as geometry diagrams, charts, tables, and scientific figures, making it suitable for evaluating whether models can solve mathematical problems grounded in visual information.
    \item \textbf{MathVision}~\citep{wang2024measuring} focuses on challenging multimodal mathematical reasoning problems. 
Its examples require models to interpret visual content and perform multi-step mathematical inference, providing a strong testbed for advanced visual-symbolic reasoning.
    \item \textbf{We-Math}~\citep{qiao2024we} provides a diagnostic evaluation for multimodal mathematical reasoning. 
By organizing problems around mathematical knowledge concepts and reasoning steps, it enables fine-grained analysis of a model's strengths and weaknesses beyond aggregate accuracy.
    \item \textbf{MathVerse}~\citep{zhang2024mathverse} is designed to examine whether multimodal models genuinely rely on diagrams in visual math problems. 
It presents problems under different visual-textual information distributions, allowing evaluation of a model's dependence on visual evidence rather than textual shortcuts.
    \item \textbf{DynaMath}~\citep{zou2025dynamath} evaluates the robustness and generalization of multimodal mathematical reasoning. 
It introduces dynamic variations of visual math problems, testing whether models can maintain consistent reasoning under controlled changes in problem instances.
    \item \textbf{MMMU-Pro}~\citep{2024arXiv240902813Y} is a challenging benchmark for multidisciplinary multimodal understanding. 
It reduces shortcut solutions from textual clues and requires models to integrate visual and textual information across diverse academic subjects.
    \item \textbf{MM-Vet}~\citep{yu2024mm} evaluates integrated multimodal capabilities through open-ended visual question answering. 
It tests whether models can jointly use perception, knowledge, reasoning, and answer generation to solve visually grounded tasks.
    \item \textbf{LogicVista}~\citep{xiao2024logicvistamultimodalllmlogical} focuses on logical reasoning in visual contexts. 
It evaluates whether models can perform structured reasoning over diagrams and other visual inputs, covering capabilities beyond purely mathematical problem solving.
    \item \textbf{NaturalBench}~\citep{li2024naturalbench} evaluates multimodal robustness under natural adversarial examples. 
It is particularly relevant for testing whether models rely on genuine visual evidence rather than language priors or superficial image-text correlations.
\end{itemize}

\paragraph{Decoding configuration.}
Each evaluation run generates one response per question
using greedy decoding with temperature 0, top-$p$ 1.0,
seed 0, and repetition penalty 1.0.
The maximum response length is 2,048 tokens for
Qwen2.5-VL models and 4,096 for Qwen3-VL models.
These backbone-specific limits are shared by the
corresponding baselines and ablation variants;
PRCO retains its published stage-specific configuration.
For each checkpoint, we conduct four separate evaluation
runs under the same settings and average the resulting
benchmark scores.
No majority voting or answer selection across runs is used.

\paragraph{Prompting protocol.}
We first construct the benchmark-specific input using
VLMEvalKit's \texttt{dataset.build\_prompt(row)}.
The original Qwen checkpoints use this input with their
native chat templates.
For GRPO, DAPO, SRPO, and the SRPO ablations, we append
the shared structured instruction in
Appendix~\ref{app:prompt_templates}, requiring a perception
segment, a reasoning segment, and a boxed final answer.
The benchmark content and appended instruction are placed
in the user message without an additional system message.
External baselines use their published evaluation prompts
and method-specific inference procedures.
Thus, prompting is matched across the controlled
GRPO, DAPO, and SRPO comparisons, while external methods
retain their respective prompting protocols.

\paragraph{Metrics and LLM-as-a-judge evaluation.}
We report the benchmark metrics listed in
Table~\ref{tab:math-benchmarks-vlmevalkit}.
Math and General are the unweighted means over the five
math-related and four general-task benchmarks,
respectively; Avg.\ is the unweighted mean over all nine.
Scores are not weighted by benchmark size.
We-Math uses the Strict score, MM-Vet permits partial
credit, DynaMath uses Overall under the Average setting,
and NaturalBench uses per-question accuracy.

For evaluation steps requiring an LLM judge, we use deepseek-v4-flash with temperature 0 and thinking=false, following the benchmark-specific
VLMEvalKit scoring rules.
Judge output limits are configured separately for each
benchmark.
NaturalBench uses rule-based scoring without an LLM judge.
The answer-evaluation judge is distinct from the
Gemini and Luna judges used in the perception-quality audit.

\subsection{Prompt Templates}
\label{app:prompt_templates}

We use the same structured template for training and
evaluating GRPO, DAPO, SRPO, and the SRPO ablations.
The template is implemented in
\texttt{perception\_reasoning\_answer.jinja}.
During training, its content field contains the training
question; during evaluation, it contains the original
benchmark input constructed by VLMEvalKit, including
the question, options, and dataset-specific instructions.
Images are supplied with the input, and the text is placed
in the user message using the model's native chat template,
without an additional system message.
GRPO and DAPO optimize their respective objectives under
this shared format, while SRPO additionally uses the
perception and reasoning spans for role-aware credit
assignment.
No ground-truth rationales or intermediate perception
annotations are used.

\begin{tcolorbox}[
    enhanced,
    breakable,
    colback=white,
    colframe=black,
    boxrule=1.0pt,
    arc=2pt,
    left=7pt,
    right=7pt,
    top=10pt,
    bottom=7pt,
    title=\textbf{Prompt for Main Rollout and Inference},
    coltitle=white,
    colbacktitle=black,
    fonttitle=\bfseries,
    attach boxed title to top left={xshift=8pt,yshift=-2pt},
    boxed title style={
        colback=black,
        colframe=black,
        arc=2pt,
        boxrule=0pt,
        left=6pt,
        right=6pt,
        top=2pt,
        bottom=2pt
    }
]
\footnotesize

\texttt{\{Question\}}

\vspace{1mm}

Analyze the image/video and solve the problem with an explicit decomposition.

\vspace{1mm}

First, write a concise but self-contained perception block that captures only the visual facts needed for solving the problem, wrapped in \texttt{<perception>} and \texttt{</perception>} tags.

\vspace{1mm}

Second, write the reasoning process grounded in the image/video and the perception block, wrapped in \texttt{<reasoning>} and \texttt{</reasoning>} tags.

\vspace{1mm}

Finally, give the answer in \texttt{\textbackslash boxed\{\}}.

\vspace{1mm}

\textbf{Required format:}

\vspace{1mm}

\texttt{<perception> visual evidence here </perception>}

\vspace{1mm}

\texttt{<reasoning> grounded reasoning here </reasoning>}

\vspace{1mm}

\texttt{\textbackslash boxed\{FINAL ANSWER\}}

\end{tcolorbox}

In addition to the main rollout template, SRPO uses an auxiliary question--image rescoring template to compute the reasoning-token support signal.
This auxiliary template removes the perception block and keeps only the reasoning span and final-answer format.
It is used only for stop-gradient rescoring in the rollout-versus-question-image comparison, and is not used as the primary rollout or evaluation prompt.

\begin{tcolorbox}[
    enhanced,
    breakable,
    colback=white,
    colframe=black,
    boxrule=1.0pt,
    arc=2pt,
    left=7pt,
    right=7pt,
    top=10pt,
    bottom=7pt,
    title=\textbf{Prompt for Question--Image Rescoring},
    coltitle=white,
    colbacktitle=black,
    fonttitle=\bfseries,
    attach boxed title to top left={xshift=8pt,yshift=-2pt},
    boxed title style={
        colback=black,
        colframe=black,
        arc=2pt,
        boxrule=0pt,
        left=6pt,
        right=6pt,
        top=2pt,
        bottom=2pt
    }
]
\footnotesize

\texttt{\{Question\}}

\vspace{1mm}

Analyze the image/video and solve the problem.

\vspace{1mm}

Write the reasoning process grounded in the image/video, wrapped in \texttt{<reasoning>} and \texttt{</reasoning>} tags.

\vspace{1mm}

Finally, give the answer in \texttt{\textbackslash boxed\{\}}.

\vspace{1mm}

\textbf{Required format:}

\vspace{1mm}

\texttt{<reasoning> grounded reasoning here </reasoning>}

\vspace{1mm}

\texttt{\textbackslash boxed\{FINAL ANSWER\}}

\end{tcolorbox}


SRPO additionally uses
\texttt{qi\_only\_reasoning\_answer.jinja}
for perception-ablated rescoring.
This template omits the perception-stage instruction
and retains the reasoning and final-answer format.
The generated perception block is removed from the
conditioning context, while the original reasoning
tokens are retained and rescored without regeneration.
This auxiliary template is used only for stop-gradient
credit computation, not for rollout generation or evaluation.

\section{Additional Case Study}
\label{app:additional_case_study}
\vspace{-2mm}

We provide qualitative comparisons between SRPO and the base Qwen3-VL-8B model across nine multimodal reasoning benchmarks. Each example presents the visual question, reference answer, and selected passages from both responses to highlight how visual evidence supports subsequent reasoning. The original wording is preserved, with $[\ldots]$ indicating omitted passages. Purple highlights perception content and orange highlights reasoning content. We organize the passages by these roles for comparison and discuss the key differences separately below each example. Both models use the same decoding settings, with the default benchmark prompt for Base and the training template for SRPO.
\setlength{\dblfloatsep}{8pt}
\setlength{\dbltextfloatsep}{8pt}
\makeatletter
\setlength{\@dblfptop}{0pt}
\setlength{\@dblfpsep}{8pt}
\setlength{\@dblfpbot}{0pt plus 1fil}
\makeatother
\definecolor{csPerc}{HTML}{7046A0}
\definecolor{csReas}{HTML}{99530D}
\definecolor{csBad}{HTML}{B02B32}
\definecolor{csGood}{HTML}{20734B}
\definecolor{csInk}{HTML}{263449}
\newcommand{\csHard}[1]{\textcolor{csBad}{#1}}
\newcommand{\csGood}[1]{\textcolor{csGood}{#1}}
\newcommand{\csPair}[2]{\noindent\begin{tabularx}{\linewidth}{@{}>{\raggedright\arraybackslash}X|>{\raggedright\arraybackslash}X@{}}#1&#2\end{tabularx}\par}
\newcommand{\csStage}[2]{\par\vspace{.7mm}{\sffamily\bfseries\color{#1}#2}\par\vspace{.3mm}}
\begin{figure*}[!p]
\centering\begin{minipage}{\textwidth}
\begin{tcolorbox}[colback=white,colframe=csInk!45,boxrule=.45pt,arc=1mm,left=2mm,right=2mm,top=1mm,bottom=1mm,before skip=0pt,after skip=0pt,title={DynaMath / 7231 --- Including an isolated vertex in graph connectivity},colbacktitle=csInk!7,coltitle=csInk,fonttitle=\small\bfseries]
\fontsize{8.5}{9.5}\selectfont\setlength{\parskip}{1pt}\setlength{\tabcolsep}{2.5mm}\setlength{\abovedisplayskip}{3pt}\setlength{\belowdisplayskip}{3pt}
\begin{minipage}[c]{.30\linewidth}\centering\includegraphics[width=\linewidth,height=22mm,keepaspectratio]{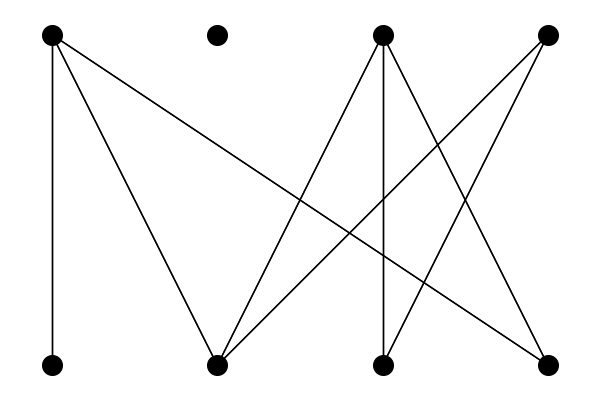}\end{minipage}\hfill\begin{minipage}[c]{.67\linewidth}\textbf{Question.} How many connected components does the graph have?

\textbf{Reference.} $2$\end{minipage}\par\vspace{1mm}
\csPair{\sffamily\bfseries Base}{\sffamily\bfseries SRPO}
\csStage{csPerc}{Perception}
\begingroup\color{csPerc}
\csPair{\textcolor{black!55}{[\ldots]} 1. Observe the graph: There are 7 vertices and several edges connecting them. \textcolor{black!55}{[\ldots]}}{\textcolor{black!55}{[\ldots]} There is one isolated vertex at the top center, not connected to any other vertex. The remaining vertices are interconnected in a complex manner, forming a single connected component. \textcolor{black!55}{[\ldots]}}
\endgroup
\csStage{csReas}{Reasoning}
\begingroup\color{csReas}
\csPair{\textcolor{black!55}{[\ldots]} 3. Since all vertices are reachable from each other, the entire graph forms a single connected component. \textcolor{black!55}{[\ldots]}}{\textcolor{black!55}{[\ldots]} 2. Count the components: - Isolated vertex: 1 component. - Connected group of other vertices: 1 component. Total components = 1 + 1 = 2. \textcolor{black!55}{[\ldots]}}
\endgroup
\csStage{csInk}{Answer}
\begingroup\color{csInk}
\csPair{\textcolor{black!55}{[\ldots]} "short answer": "1.000" \textcolor{black!55}{[\ldots]}}{\textcolor{black!55}{[\ldots]} $\boxed{2.0}$ \textcolor{black!55}{[\ldots]}}
\endgroup
\par\vspace{1mm}\textbf{Our analysis.} The graph contains eight marked vertices. The edge-connected portion contains seven. The remaining top vertex is isolated. SRPO includes this visually small but topologically decisive object, while the base traversal description omits it. The difference is substantive connectivity, not JSON versus boxed-answer formatting.
\end{tcolorbox}
\par\vspace{3mm}
\begin{tcolorbox}[colback=white,colframe=csInk!45,boxrule=.45pt,arc=1mm,left=2mm,right=2mm,top=1mm,bottom=1mm,before skip=0pt,after skip=0pt,title={NaturalBenchDataset / 6985 --- Recognizing the objects before inferring the action},colbacktitle=csInk!7,coltitle=csInk,fonttitle=\small\bfseries]
\fontsize{8.5}{9.5}\selectfont\setlength{\parskip}{1pt}\setlength{\tabcolsep}{2.5mm}\setlength{\abovedisplayskip}{3pt}\setlength{\belowdisplayskip}{3pt}
\begin{minipage}[c]{.30\linewidth}\centering\includegraphics[width=\linewidth,height=22mm,keepaspectratio]{figures/cases/NaturalBenchDataset_6985_0.jpg}\end{minipage}\hfill\begin{minipage}[c]{.67\linewidth}\textbf{Question.} What activity is the woman engaging in? A: The woman is standing with a man. B: The woman is trying to light a cigarette.

\textbf{Reference.} B.\end{minipage}\par\vspace{1mm}
\csPair{\sffamily\bfseries Base}{\sffamily\bfseries SRPO}
\csStage{csPerc}{Perception}
\begingroup\color{csPerc}
\csPair{[\ldots] The woman is looking at her phone, likely texting or using an app. [\ldots]}{[\ldots] The woman in the image is holding a small object that appears to be a cigarette and a lighter, and she is looking down at them with a focused expression, suggesting she is in the process of lighting the cigarette. [\ldots]}
\endgroup
\csStage{csReas}{Reasoning}
\begingroup\color{csReas}
\csPair{[\ldots] None of the given options describe this activity. [\ldots]}{[\ldots] The woman is holding a cigarette and a lighter, and her posture and gaze indicate she is actively attempting to light the cigarette. Neither a man nor any other person is visible with her, eliminating option A. The visual evidence strongly supports option B. [\ldots]}
\endgroup
\csStage{csInk}{Answer}
\begingroup\color{csInk}
\csPair{[\ldots] Neither option is correct. [\ldots]}{[\ldots] $\boxed{B}$ [\ldots]}
\endgroup
\par\vspace{1mm}\textbf{Our analysis.} The base response identifies a phone and rejects both options. SRPO identifies a cigarette and lighter and selects the matching action. The contrast is object recognition followed by action inference.
\end{tcolorbox}
\end{minipage}\end{figure*}

\begin{figure*}[!p]
\centering\begin{minipage}{\textwidth}
\begin{tcolorbox}[colback=white,colframe=csInk!45,boxrule=.45pt,arc=1mm,left=2mm,right=2mm,top=1mm,bottom=1mm,before skip=0pt,after skip=0pt,title={MathVista\_MINI / 111 --- Attribute binding before subtraction},colbacktitle=csInk!7,coltitle=csInk,fonttitle=\small\bfseries]
\fontsize{8.5}{9.5}\selectfont\setlength{\parskip}{1pt}\setlength{\tabcolsep}{2.5mm}\setlength{\abovedisplayskip}{3pt}\setlength{\belowdisplayskip}{3pt}
\begin{minipage}[c]{.30\linewidth}\centering\includegraphics[width=\linewidth,height=22mm,keepaspectratio]{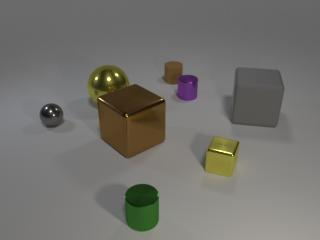}\end{minipage}\hfill\begin{minipage}[c]{.67\linewidth}\textbf{Question.} Subtract all big gray matte objects and all small metallic cylinders. How many objects remain?

\textbf{Reference.} $5$\end{minipage}\par\vspace{1mm}
\csPair{\sffamily\bfseries Base}{\sffamily\bfseries SRPO}
\csStage{csPerc}{Perception}
\begingroup\color{csPerc}
\csPair{\textcolor{black!55}{[\ldots]} 6. Small brown metallic cylinder (behind the purple one) \textcolor{black!55}{[\ldots]}}{\textcolor{black!55}{[\ldots]} The image contains the following objects: a small gray metallic sphere, a large gold metallic sphere, a large bronze metallic cube, a small green metallic cylinder, a small yellow metallic cube, a large gray matte cube, a small brown matte cylinder, and a small purple metallic cylinder. \textcolor{black!55}{[\ldots]}}
\endgroup
\csStage{csReas}{Reasoning}
\begingroup\color{csReas}
\csPair{\textcolor{black!55}{[\ldots]} Total subtracted: 1 (gray cube) + 3 (small metallic cylinders) = 4 objects.\par \par Remaining objects: 8 - 4 = 4. \textcolor{black!55}{[\ldots]}}{\textcolor{black!55}{[\ldots]} Next, subtract all big gray matte things: The large gray matte cube qualifies. Remove it. Remaining: 7 objects. Next, subtract all small metallic cylinders: The small green metallic cylinder and the small purple metallic cylinder qualify. Remove them. Remaining: 7 - 2 = 5 objects. \textcolor{black!55}{[\ldots]}}
\endgroup
\csStage{csInk}{Answer}
\begingroup\color{csInk}
\csPair{\textcolor{black!55}{[\ldots]} Therefore, there are 4 objects left. \textcolor{black!55}{[\ldots]}}{\textcolor{black!55}{[\ldots]} $\boxed{5}$ \textcolor{black!55}{[\ldots]}}
\endgroup
\par\vspace{1mm}\textbf{Our analysis.} Both responses count eight objects and perform consistent subtraction. The decisive difference is the material assigned to one cylinder. SRPO preserves the brown matte cylinder, so its intermediate object description supports the correct removal set.
\end{tcolorbox}
\par\vspace{3mm}
\begin{tcolorbox}[colback=white,colframe=csInk!45,boxrule=.45pt,arc=1mm,left=2mm,right=2mm,top=1mm,bottom=1mm,before skip=0pt,after skip=0pt,title={LogicVista / 331 --- Keeping aggregation consistent with the extracted chart values},colbacktitle=csInk!7,coltitle=csInk,fonttitle=\small\bfseries]
\fontsize{8.5}{9.5}\selectfont\setlength{\parskip}{1pt}\setlength{\tabcolsep}{2.5mm}\setlength{\abovedisplayskip}{3pt}\setlength{\belowdisplayskip}{3pt}
\begin{minipage}[c]{.30\linewidth}\centering\includegraphics[width=\linewidth,height=22mm,keepaspectratio]{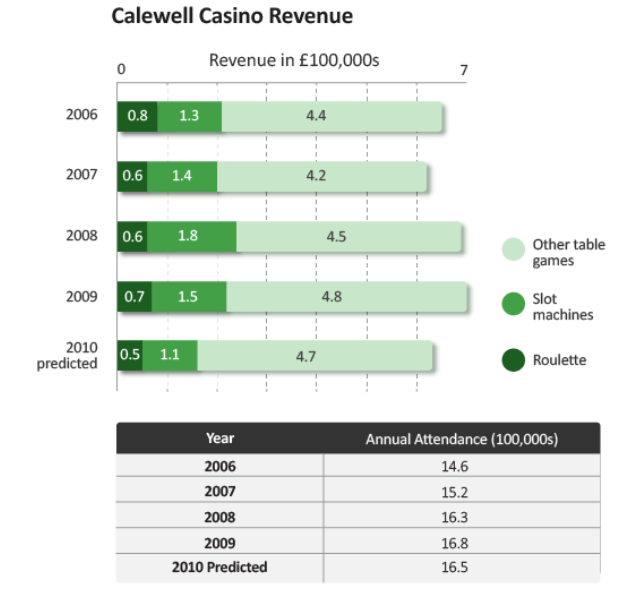}\end{minipage}\hfill\begin{minipage}[c]{.67\linewidth}\textbf{Question.} How much did combined Slot machines and Roulette revenue differ from Other table games in 2006--2009, in millions of pounds? Options: A: $0.9$; B: $9.0$; C: $9.2$; D: $0.92$; E: none.

\textbf{Reference.} D: \pounds$0.92$ million.\end{minipage}\par\vspace{1mm}
\csPair{\sffamily\bfseries Base}{\sffamily\bfseries SRPO}
\csStage{csPerc}{Perception}
\begingroup\color{csPerc}
\csPair{\textcolor{black!55}{[\ldots]} The chart shows revenue in £100,000s. So, we will work in those units and convert to £millions at the end (since 1 £100,000 = 0.1 £million). \textcolor{black!55}{[\ldots]}}{\textcolor{black!55}{[\ldots]} The image displays a stacked bar chart titled "Calewell Casino Revenue" showing revenue in £100,000s for the years 2006-2009 and a predicted value for 2010. \textcolor{black!55}{[\ldots]}}
\endgroup
\csStage{csReas}{Reasoning}
\begingroup\color{csReas}
\csPair{\textcolor{black!55}{[\ldots]} Total difference = 2.3 + 2.2 + 2.1 + 2.6 = 11.2\par \par This is in £100,000s.\par \par Convert to £millions: 11.2 \ensuremath{\times} 0.1 = 1.12 million \textcolor{black!55}{[\ldots]}}{\textcolor{black!55}{[\ldots]} 2.1 + 2.0 + 2.4 + 2.2 = 8.7 (in £100,000s) \textcolor{black!55}{[\ldots]} 4.4 + 4.2 + 4.5 + 4.8 = 17.9 (in £100,000s) 5. Calculate the difference: Other table games revenue - (Slot machines + Roulette) revenue = 17.9 - 8.7 = 9.2 (in £100,000s) 6. Convert to £millions: 9.2 £100,000s = 9.2 * 0.1 = 0.92 £millions. 7. Match with options: (D) 0.92 \textcolor{black!55}{[\ldots]}}
\endgroup
\csStage{csInk}{Answer}
\begingroup\color{csInk}
\csPair{\textcolor{black!55}{[\ldots]} (E) None of these \textcolor{black!55}{[\ldots]}}{\textcolor{black!55}{[\ldots]} $\boxed{D}$ \textcolor{black!55}{[\ldots]}}
\endgroup
\par\vspace{1mm}\textbf{Our analysis.} This is an aggregation error, not a unit-reading error: both responses identify the correct chart values and conversion factor. The base response makes the checkable arithmetic mistake $2.3+2.2+2.1+2.6=11.2$. SRPO obtains the correct difference by aggregating category totals before subtraction.
\end{tcolorbox}
\end{minipage}\end{figure*}

\begin{figure*}[!p]
\centering\begin{minipage}{\textwidth}
\begin{tcolorbox}[colback=white,colframe=csInk!45,boxrule=.45pt,arc=1mm,left=2mm,right=2mm,top=1mm,bottom=1mm,before skip=0pt,after skip=0pt,title={MathVision\_MINI / 1060 --- Testing the interior pattern instead of inheriting square symmetry},colbacktitle=csInk!7,coltitle=csInk,fonttitle=\small\bfseries]
\fontsize{8.5}{9.5}\selectfont\setlength{\parskip}{1pt}\setlength{\tabcolsep}{2.5mm}\setlength{\abovedisplayskip}{3pt}\setlength{\belowdisplayskip}{3pt}
\begin{minipage}[c]{.30\linewidth}\centering\includegraphics[width=\linewidth,height=22mm,keepaspectratio]{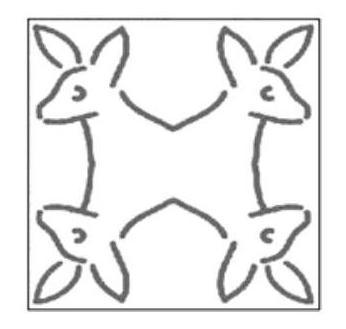}\end{minipage}\hfill\begin{minipage}[c]{.67\linewidth}\textbf{Question.} How many lines of symmetry does the figure have?

\textbf{Reference.} $2$\end{minipage}\par\vspace{1mm}
\csPair{\sffamily\bfseries Base}{\sffamily\bfseries SRPO}
\csStage{csPerc}{Perception}
\begingroup\color{csPerc}
\csPair{\textcolor{black!55}{[\ldots]} The figure is a square with four identical, stylized animal-like shapes (resembling deer or rabbits) arranged symmetrically: one at each corner of the square. \textcolor{black!55}{[\ldots]}}{\textcolor{black!55}{[\ldots]} The image displays a square containing a symmetrical pattern of four deer-like figures. Two deer faces are positioned at the top-left and top-right corners, facing inward toward the center. Two deer bodies are positioned at the bottom-left and bottom-right corners, also facing inward. The overall design forms a cross-like symmetry within the square. \textcolor{black!55}{[\ldots]}}
\endgroup
\csStage{csReas}{Reasoning}
\begingroup\color{csReas}
\csPair{\textcolor{black!55}{[\ldots]} All four lines (vertical, horizontal, and two diagonals) divide the figure into two identical, mirror-image halves. \textcolor{black!55}{[\ldots]}}{\textcolor{black!55}{[\ldots]} 3. Diagonal Lines of Symmetry: Checking the diagonals, the top-left to bottom-right diagonal does not produce a mirror image (the top-left deer's orientation and position do not match the bottom-right deer's). Similarly, the top-right to bottom-left diagonal does not produce a mirror image. Therefore, only the vertical and horizontal lines are lines of symmetry. This gives us exactly 2 lines of symmetry. \textcolor{black!55}{[\ldots]}}
\endgroup
\csStage{csInk}{Answer}
\begingroup\color{csInk}
\csPair{\textcolor{black!55}{[\ldots]} Therefore, the figure has 4 lines of symmetry. \textcolor{black!55}{[\ldots]}}{\textcolor{black!55}{[\ldots]} $\boxed{2}$ \textcolor{black!55}{[\ldots]}}
\endgroup
\par\vspace{1mm}\textbf{Our analysis.} The outer square alone has four reflection axes, but the complete figure does not. SRPO checks the internal motif arrangement and retains only the two valid axes. The base response explicitly accepts diagonal reflections that the image does not support.
\end{tcolorbox}
\par\vspace{3mm}
\begin{tcolorbox}[colback=white,colframe=csInk!45,boxrule=.45pt,arc=1mm,left=2mm,right=2mm,top=1mm,bottom=1mm,before skip=0pt,after skip=0pt,title={MathVerse\_MINI\_Vision\_Only / 2190 --- Carrying the derived parameter into the final equation},colbacktitle=csInk!7,coltitle=csInk,fonttitle=\small\bfseries]
\fontsize{8.5}{9.5}\selectfont\setlength{\parskip}{1pt}\setlength{\tabcolsep}{2.5mm}\setlength{\abovedisplayskip}{3pt}\setlength{\belowdisplayskip}{3pt}
\begin{minipage}[c]{.30\linewidth}\centering\includegraphics[width=\linewidth,height=22mm,keepaspectratio]{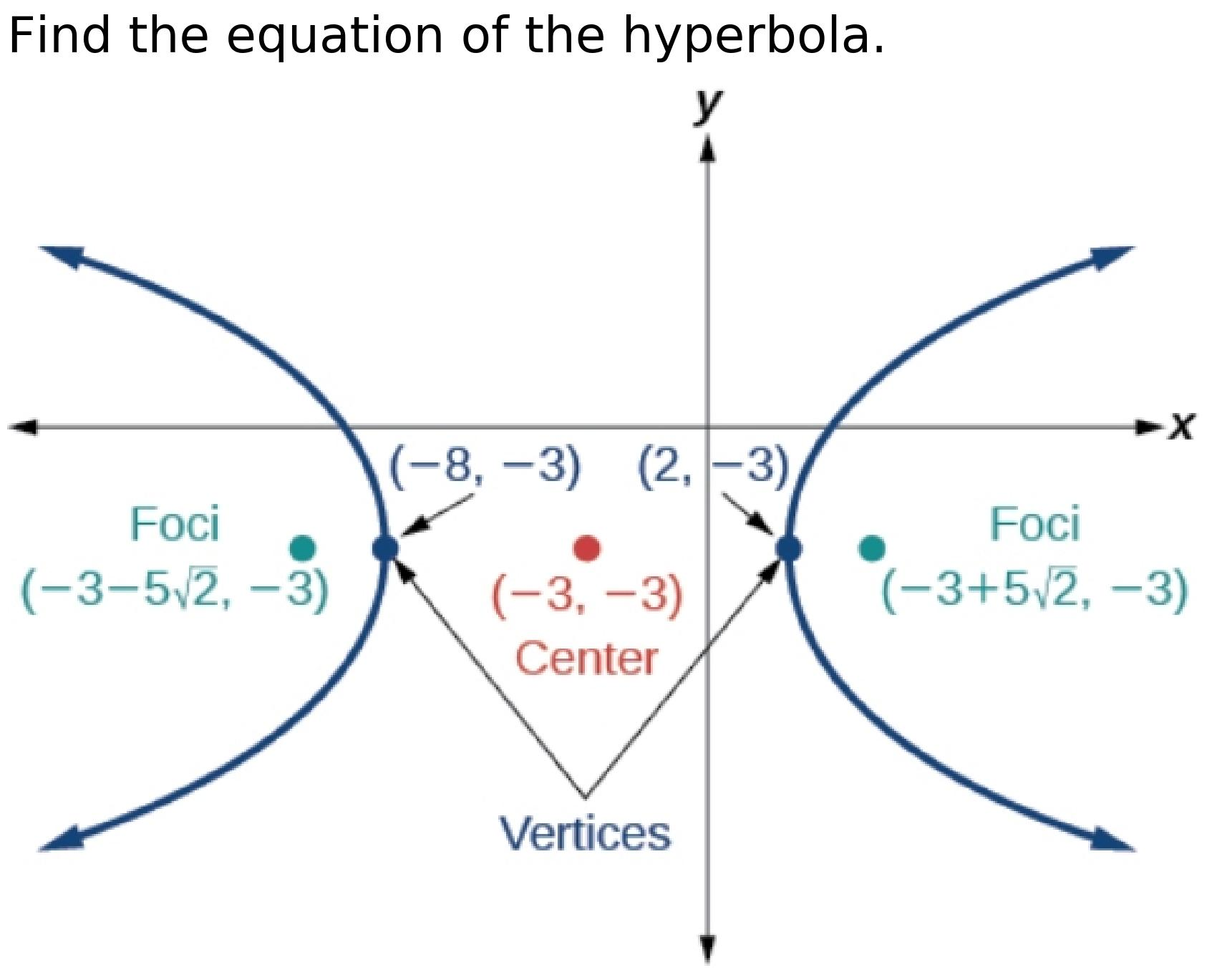}\end{minipage}\hfill\begin{minipage}[c]{.67\linewidth}\textbf{Question.} Find the equation of the horizontal hyperbola shown in the image. Its center is $(-3,-3)$, vertices are $(-8,-3)$ and $(2,-3)$, and foci are $(-3\pm5\sqrt{2},-3)$.

\textbf{Reference.} $\dfrac{(x+3)^2}{25}-\dfrac{(y+3)^2}{25}=1$\end{minipage}\par\vspace{1mm}
\csPair{\sffamily\bfseries Base}{\sffamily\bfseries SRPO}
\csStage{csPerc}{Perception}
\begingroup\color{csPerc}
\csPair{\textcolor{black!55}{[\ldots]} This is derived from the given center $(-3, -3)$, vertices at $(-8, -3)$ and $(2, -3)$, and foci at $(-3 \pm 5\sqrt{2}, -3)$. The transverse axis is horizontal, \textcolor{black!55}{[\ldots]}}{\textcolor{black!55}{[\ldots]} The image displays a hyperbola with its center at (-3, -3). The vertices are at (-8, -3) and (2, -3), and the foci are at (-3-5\ensuremath{\surd}2, -3) and (-3+5\ensuremath{\surd}2, -3). The hyperbola opens horizontally, as indicated by the positions of the vertices and foci along the horizontal line y = -3. \textcolor{black!55}{[\ldots]}}
\endgroup
\csStage{csReas}{Reasoning}
\begingroup\color{csReas}
\csPair{\textcolor{black!55}{[\ldots]} From the vertices, $a = 5$, and from the foci, $c = 5\sqrt{2}$. Then $b^2 = c^2 - a^2 = 50 - 25 = 25$. \textcolor{black!55}{[\ldots]}}{\textcolor{black!55}{[\ldots]} 4. For a hyperbola, the relationship is \(c^2 = a^2 + b^2\). Substituting the known values: \(50 = 25 + b^2\), so \(b^2 = 25\). \textcolor{black!55}{[\ldots]}}
\endgroup
\csStage{csInk}{Answer}
\begingroup\color{csInk}
\csPair{\textcolor{black!55}{[\ldots]} Final Answer: $\frac{(x + 3)^2}{25} - \frac{(y + 3)^2}{50} = 1$ \textcolor{black!55}{[\ldots]}}{\textcolor{black!55}{[\ldots]} $\boxed{\frac{(x+3)^2}{25} - \frac{(y+3)^2}{25} = 1}$ \textcolor{black!55}{[\ldots]}}
\endgroup
\par\vspace{1mm}\textbf{Our analysis.} Visual extraction succeeds in both responses. The failure occurs at the reasoning-to-answer transition: the base response replaces $b^2$ with $c^2$ in its final equation despite computing $b^2$ correctly. SRPO retains the derived value through the final substitution.
\end{tcolorbox}
\end{minipage}\end{figure*}

\begin{figure*}[!p]
\centering\begin{minipage}{\textwidth}
\begin{tcolorbox}[colback=white,colframe=csInk!45,boxrule=.45pt,arc=1mm,left=2mm,right=2mm,top=1mm,bottom=1mm,before skip=0pt,after skip=0pt,title={WeMath / 47 --- Binding a dimension label to the radius},colbacktitle=csInk!7,coltitle=csInk,fonttitle=\small\bfseries]
\fontsize{8.5}{9.5}\selectfont\setlength{\parskip}{1pt}\setlength{\tabcolsep}{2.5mm}\setlength{\abovedisplayskip}{3pt}\setlength{\belowdisplayskip}{3pt}
\begin{minipage}[c]{.30\linewidth}\centering\includegraphics[width=\linewidth,height=22mm,keepaspectratio]{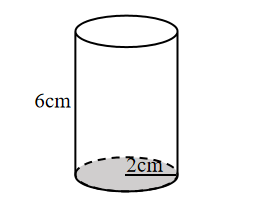}\end{minipage}\hfill\begin{minipage}[c]{.67\linewidth}\textbf{Question.} Find the total surface area of the cylinder. Options: A: $24\pi$; B: $16\pi$; C: $32\pi$; D: no correct answer.

\textbf{Reference.} C: $32\pi\,\mathrm{cm}^2$\end{minipage}\par\vspace{1mm}
\csPair{\sffamily\bfseries Base}{\sffamily\bfseries SRPO}
\csStage{csPerc}{Perception}
\begingroup\color{csPerc}
\csPair{\textcolor{black!55}{[\ldots]} - The radius r = 2 cm (given as the diameter of the base, so radius is half of that, but wait — the figure shows "2cm" as a line across the base, which is the diameter, not the radius. Let me re-read. The label "2cm" is drawn across the base, indicating the diameter. So, diameter = 2 cm, therefore radius r = 1 cm.\par - The height h = 6 cm. \textcolor{black!55}{[\ldots]}}{\textcolor{black!55}{[\ldots]} The image shows a cylinder with a height labeled as 6 cm and a radius labeled as 2 cm at the base. \textcolor{black!55}{[\ldots]}}
\endgroup
\csStage{csReas}{Reasoning}
\begingroup\color{csReas}
\csPair{\textcolor{black!55}{[\ldots]} - Area of one base = \ensuremath{\pi}r\textsuperscript{2} = \ensuremath{\pi}(1)\textsuperscript{2} = \ensuremath{\pi} cm\textsuperscript{2}\par - Area of two bases = 2\ensuremath{\pi} cm\textsuperscript{2}\par - Lateral surface area = 2\ensuremath{\pi}rh = 2\ensuremath{\pi}(1)(6) = 12\ensuremath{\pi} cm\textsuperscript{2}\par - Total surface area = 2\ensuremath{\pi} + 12\ensuremath{\pi} = 14\ensuremath{\pi} cm\textsuperscript{2} \textcolor{black!55}{[\ldots]}}{\textcolor{black!55}{[\ldots]} First, calculate the area of the two bases: 2 \ensuremath{\times} \ensuremath{\pi} \ensuremath{\times} r\textsuperscript{2} = 2 \ensuremath{\times} \ensuremath{\pi} \ensuremath{\times} 2\textsuperscript{2} = 2 \ensuremath{\times} \ensuremath{\pi} \ensuremath{\times} 4 = 8\ensuremath{\pi} cm\textsuperscript{2}. Next, calculate the lateral surface area: 2\ensuremath{\pi}r \ensuremath{\times} h = 2\ensuremath{\pi} \ensuremath{\times} 2 \ensuremath{\times} 6 = 24\ensuremath{\pi} cm\textsuperscript{2}. Finally, add them together: Total Surface Area = 8\ensuremath{\pi} + 24\ensuremath{\pi} = 32\ensuremath{\pi} cm\textsuperscript{2}. \textcolor{black!55}{[\ldots]}}
\endgroup
\csStage{csInk}{Answer}
\begingroup\color{csInk}
\csPair{\textcolor{black!55}{[\ldots]} 14\ensuremath{\pi} is not among the options. Therefore, the correct answer is D. No correct answer. \textcolor{black!55}{[\ldots]}}{\textcolor{black!55}{[\ldots]} $\boxed{C}$ \textcolor{black!55}{[\ldots]}}
\endgroup
\par\vspace{1mm}\textbf{Our analysis.} The formula is not the bottleneck: both models use it correctly. Binding the label to the radius rather than the diameter changes both terms of the computation. This panel displays subquestion 47 from group \texttt{3stpes\_12}. The full group and the displayed subquestion are scored separately in the evidence files. Both models answer the group's final composite question correctly, but the base model fails this subquestion and therefore fails strict group mastery.
\end{tcolorbox}
\par\vspace{3mm}
\begin{tcolorbox}[colback=white,colframe=csInk!45,boxrule=.45pt,arc=1mm,left=2mm,right=2mm,top=1mm,bottom=1mm,before skip=0pt,after skip=0pt,title={MMVet / 35 --- Following the complete infographic sequence across rows},colbacktitle=csInk!7,coltitle=csInk,fonttitle=\small\bfseries]
\fontsize{8.5}{9.5}\selectfont\setlength{\parskip}{1pt}\setlength{\tabcolsep}{2.5mm}\setlength{\abovedisplayskip}{3pt}\setlength{\belowdisplayskip}{3pt}
\begin{minipage}[c]{.30\linewidth}\centering\includegraphics[width=\linewidth,height=22mm,keepaspectratio]{figures/cases/MMVet_35_0.png}\end{minipage}\hfill\begin{minipage}[c]{.67\linewidth}\textbf{Question.} What should I do before cutting herbs, sausage, and mushrooms?

\textbf{Reference.} Tilt pan slightly.\end{minipage}\par\vspace{1mm}
\csPair{\sffamily\bfseries Base}{\sffamily\bfseries SRPO}
\csStage{csPerc}{Perception}
\begingroup\color{csPerc}
\csPair{\textcolor{black!55}{[\ldots]} 1.  Beat eggs with salt (This is the first step shown in the top-left circle).\par 2.  Cut herbs, sausage, and mushrooms (This is the second step shown in the bottom-left circle). \textcolor{black!55}{[\ldots]}}{\textcolor{black!55}{[\ldots]} The first step shows beating eggs with salt. The second step shows heating butter in a frying pan. The third step shows pouring eggs into the pan. The fourth step shows tilting the pan slightly. The fifth step shows cutting herbs, sausage, and mushrooms. \textcolor{black!55}{[\ldots]}}
\endgroup
\csStage{csReas}{Reasoning}
\begingroup\color{csReas}
\csPair{\textcolor{black!55}{[\ldots]} The infographic presents the steps in a specific order, implying that preparing the eggs (beating them with salt) is a prerequisite to preparing the other ingredients (cutting the herbs, sausage, and mushrooms). \textcolor{black!55}{[\ldots]}}{\textcolor{black!55}{[\ldots]} These four steps occur before Step 5, which is "CUT HERBS, SAUSAGE AND MUSHROOMS." Therefore, before cutting the herbs, sausage, and mushrooms, you should complete the steps of beating eggs, heating butter, pouring eggs into the pan, and tilting the pan. \textcolor{black!55}{[\ldots]}}
\endgroup
\csStage{csInk}{Answer}
\begingroup\color{csInk}
\csPair{\textcolor{black!55}{[\ldots]} Therefore, the action you should do before cutting the herbs, sausage, and mushrooms is to beat the eggs with salt. \textcolor{black!55}{[\ldots]}}{\textcolor{black!55}{[\ldots]} Beat eggs with salt, heat butter in frying pan, pour eggs into pan, and tilt pan slightly \textcolor{black!55}{[\ldots]}}
\endgroup
\par\vspace{1mm}\textbf{Our analysis.} The key difference is reading order across the two-row infographic. SRPO recovers the omitted intermediate actions and includes the immediately preceding step required by the reference answer. Its response is broader than the short reference (it lists all four prior actions). The official scorer gives it full credit.
\end{tcolorbox}
\end{minipage}\end{figure*}

\begin{figure*}[!p]
\centering\begin{minipage}{\textwidth}
\begin{tcolorbox}[colback=white,colframe=csInk!45,boxrule=.45pt,arc=1mm,left=2mm,right=2mm,top=1mm,bottom=1mm,before skip=0pt,after skip=0pt,title={MMMU\_Pro\_V / 497 --- Reading a local slope before applying the product rule},colbacktitle=csInk!7,coltitle=csInk,fonttitle=\small\bfseries]
\fontsize{8.5}{9.5}\selectfont\setlength{\parskip}{1pt}\setlength{\tabcolsep}{2.5mm}\setlength{\abovedisplayskip}{3pt}\setlength{\belowdisplayskip}{3pt}
\begin{minipage}[c]{.30\linewidth}\centering\includegraphics[trim=660bp 1000bp 2180bp 385bp,clip,width=\linewidth,height=22mm,keepaspectratio]{figures/cases/MMMU_Pro_V_497_0.png}\end{minipage}\hfill\begin{minipage}[c]{.67\linewidth}\textbf{Question.} Given $h(x)=f(x)g(x)$ and the plotted functions, find $h\prime(1)$. Relevant options: D: $0$; F: $2$.

\textbf{Reference.} F: $2$\end{minipage}\par\vspace{1mm}
\csPair{\sffamily\bfseries Base}{\sffamily\bfseries SRPO}
\csStage{csPerc}{Perception}
\begingroup\color{csPerc}
\csPair{\textcolor{black!55}{[\ldots]} Since \( f(x) \) is a straight line, its derivative is constant. \textcolor{black!55}{[\ldots]} The graph of \( g(x) \) has a corner at \( x = 3 \), but at \( x = 1 \), it is part of a straight line segment.\par    - The segment from \( (0, 0) \) to \( (3, 1) \) is linear. \textcolor{black!55}{[\ldots]}}{\textcolor{black!55}{[\ldots]} The graph of f(x) is a straight line passing through points (0, 4) and (2, 2). \textcolor{black!55}{[\ldots]} The graph of g(x) at x=1 is part of a dashed line segment from (0,0) to (2,2). \textcolor{black!55}{[\ldots]}}
\endgroup
\csStage{csReas}{Reasoning}
\begingroup\color{csReas}
\csPair{\textcolor{black!55}{[\ldots]} Slope \( = \frac{1 - 0}{3 - 0} = \frac{1}{3} \). \textcolor{black!55}{[\ldots]} \(\begin{aligned}h'(1)&= f'(1)g(1) + f(1)g'(1)\\&= (-1)(1) + (3)\left(\frac{1}{3}\right)\\&= -1 + 1 = 0\end{aligned}\) \textcolor{black!55}{[\ldots]}}{\textcolor{black!55}{[\ldots]} So f'(x) = -1 for all x, including x=1. \textcolor{black!55}{[\ldots]} The slope of this segment is (2-0)/(2-0) = 2/2 = 1. So g'(1) = 1. \textcolor{black!55}{[\ldots]} Now apply the product rule: h'(1) = f'(1)g(1) + f(1)g'(1) = (-1)(1) + (3)(1) = -1 + 3 = 2. Therefore, h'(1) = 2. \textcolor{black!55}{[\ldots]}}
\endgroup
\csStage{csInk}{Answer}
\begingroup\color{csInk}
\csPair{\textcolor{black!55}{[\ldots]} Therefore, the answer is \( \boxed{0} \), which corresponds to option (D). \textcolor{black!55}{[\ldots]}}{\textcolor{black!55}{[\ldots]} $\boxed{F}$ \textcolor{black!55}{[\ldots]}}
\endgroup
\par\vspace{1mm}\textbf{Our analysis.} Both models choose the correct calculus rule. The difference is a single visual intermediate variable: the slope of the dashed segment near $x=1$. SRPO reads endpoints consistent with the plot. The base response uses a nonexistent point and propagates that error through otherwise valid algebra. Both outputs also overgeneralize the solid function as a straight line; SRPO explicitly states a global derivative of $-1$. The successful contrast here is the local dashed-line slope at $x=1$.
\end{tcolorbox}
\end{minipage}\end{figure*}

\end{CJK*}
\end{document}